\documentclass[sigconf]{acmart}
\usepackage{hyperref}
\usepackage{colortbl}
\usepackage{float}

\usepackage{graphicx}
\usepackage{xspace}
\usepackage[table]{xcolor}
\usepackage{booktabs} 
\usepackage{multirow}
\usepackage{array}
\usepackage{amsmath}
\usepackage{tcolorbox}
\usepackage{bbm}

\usepackage{ragged2e}
\usepackage{rotating} 

\newcolumntype{L}[1]{>{\RaggedRight\arraybackslash}p{#1}}
\newcolumntype{C}[1]{>{\Centering\arraybackslash}p{#1}}

\newcommand{\eg}{\emph{e.g.,}\xspace}
\newcommand{\ie}{\emph{i.e.,}\xspace}
\newcommand{\etc}{\emph{etc.}\xspace}

\definecolor{lightgrayv}{HTML}{F4F3F8} 
\definecolor{mydarkblue}{rgb}{0,0.08,0.45}
\newcommand{\baby}{\text{MineralExp}\xspace}

\AtBeginDocument{%
  }

\setcopyright{none}
\renewcommand\footnotetextcopyrightpermission[1]{}
\begin{document}

\title{\baby: Benchmarking LLM Agents for Experimental Design in Real-World Mineral Exploration}

\def\huggingface{\raisebox{-1.5pt}{\includegraphics[height=1.0em]{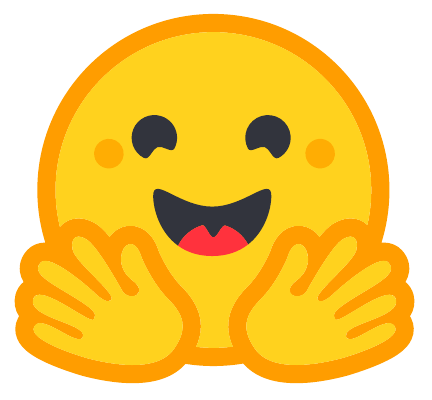}}}
\def\github{\raisebox{-1.5pt}{\includegraphics[height=1.0em]{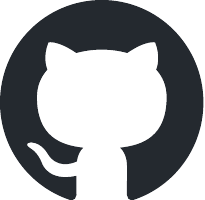}}}
\def\web{\raisebox{-1.5pt}{\includegraphics[height=1.0em]{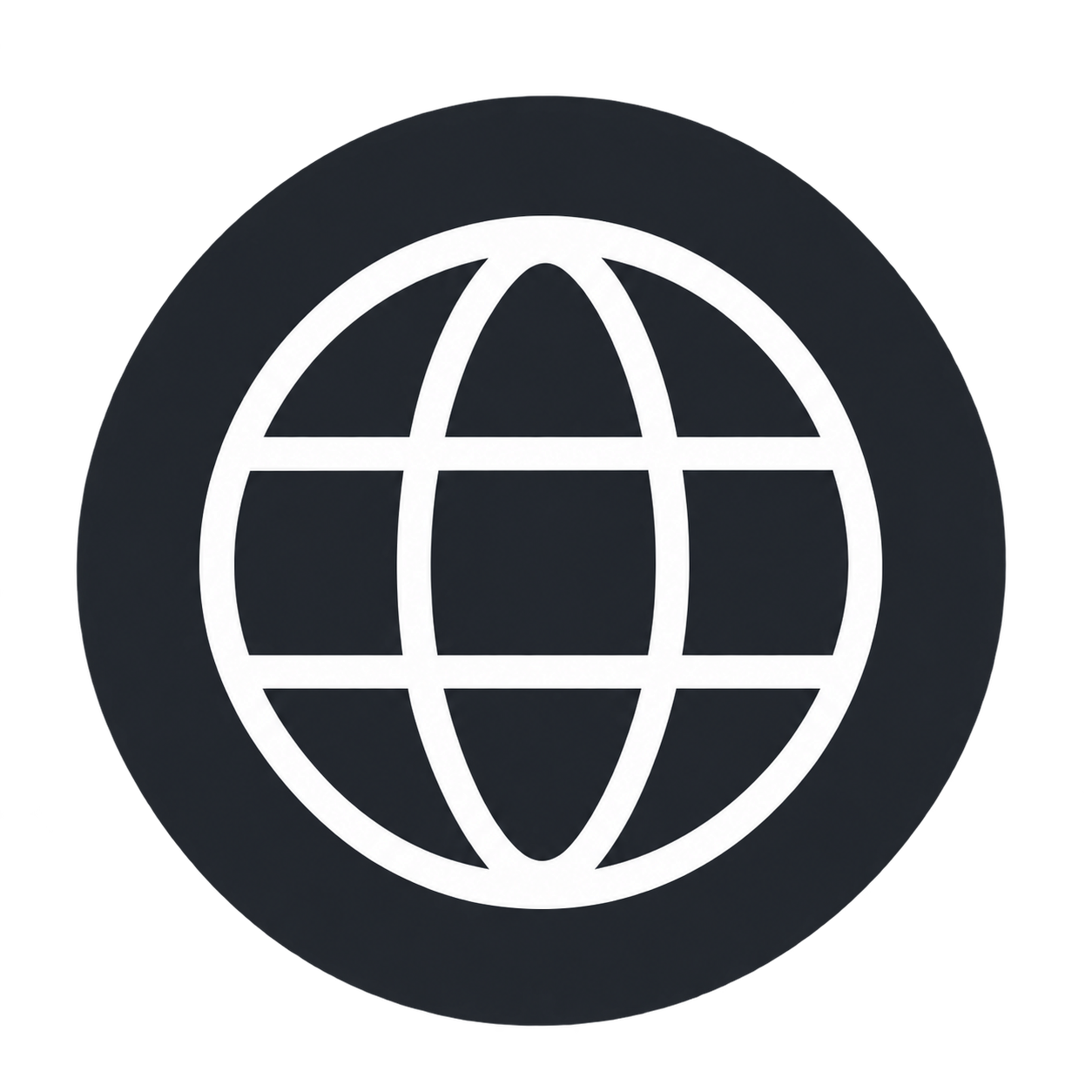}}}


\author{Bing Wang}
\authornote{Bing Wang and Mingyang Wang contributed equally to this benchmark.}
\orcid{0000-0002-1304-3718}
\affiliation{%
  \institution{College of Computer Science and Technology, and Key Laboratory of Symbolic Computation and Knowledge Engineering of the MoE, Jilin University}
  \city{Changchun}
  \country{China}}
\email{wangbing1416@gmail.com}

\author{Mingyang Wang}
\authornotemark[1]
\orcid{0000-0003-4034-6383}
\affiliation{%
  \institution{State Key Laboratory of Geological Processes and Mineral Resources, Frontiers Science Center for Deep-time Digital Earth, School of Earth Science and Resources, China University of Geosciences (Beijing)}
  \city{Beijing}
  \country{China}}
\email{Huysmanwang@qq.com}

\author{Yuanhang Mi}
\orcid{}
\author{Yue Xu}
\orcid{}
\affiliation{%
  \institution{College of Computer Science and Technology, and Key Laboratory of Symbolic Computation and Knowledge Engineering of the MoE, Jilin University}
  \city{Changchun}
  \country{China}}
\email{miyh26@mails.jlu.edu.cn}
\email{yuexu26@mails.jlu.edu.cn}

\author{Kaibo Su}
\orcid{}
\author{Li'ang Xiao}
\orcid{}
\affiliation{%
  \institution{College of Computer Science and Technology, and Key Laboratory of Symbolic Computation and Knowledge Engineering of the MoE, Jilin University}
  \city{Changchun}
  \country{China}}
\email{sukb26@mails.jlu.edu.cn}
\email{xiaola26@mails.jlu.edu.cn}

\author{Chunguo Wu}
\orcid{0000-0002-8161-3997}
\affiliation{%
  \institution{College of Computer Science and Technology, and Key Laboratory of Symbolic Computation and Knowledge Engineering of the MoE, Jilin University}
  \city{Changchun}
  \country{China}}
\email{wucg@jlu.edu.cn}

\author{Changchun Li}
\authornote{Changchun Li and Ximing Li are the corresponding authors.}
\orcid{0000-0002-8001-2655}
\author{Ximing Li}
\authornotemark[2]
\orcid{0000-0001-8190-5087}
\affiliation{%
  \institution{College of Computer Science and Technology, and Key Laboratory of Symbolic Computation and Knowledge Engineering of the MoE, Jilin University}
  \city{Changchun}
  \country{China}}
\email{changchunli93@gmail.com}
\email{liximing86@gmail.com}

\author{Kun-Feng Qiu}
\orcid{0000-0002-3185-9446}
\affiliation{%
  \institution{State Key Laboratory of Geological Processes and Mineral Resources, Frontiers Science Center for Deep-time Digital Earth, School of Earth Science and Resources, China University of Geosciences (Beijing)}
  \city{Beijing}
  \country{China}}
\email{kunfengqiu@qq.com}

\author{Bo Yang}
\orcid{0000-0003-1927-8419}
\affiliation{%
  \institution{College of Computer Science and Technology, and Key Laboratory of Symbolic Computation and Knowledge Engineering of the MoE, Jilin University}
  \city{Changchun}
  \country{China}}
\email{ybo@jlu.edu.cn}

\renewcommand{\shortauthors}{Bing Wang, Mingyang Wang, et al.}

\begin{abstract}
  
In mineral exploration, verifying geological hypotheses relies on designing a variety of geochemical experiments. While large language model (LLM) agents increasingly automate downstream geoscientific data analysis, a crucial upstream capability remains largely unexplored: \textit{automated geochemical experimental design}, \ie deciding which experiments to conduct before committing to costly laboratory work. Evaluating this capability of LLM agents is difficult for two reasons. First, geochemical experiments exhibit high methodological diversity and subjectivity: to test the same hypothesis, different geologists may choose different minerals, instruments, and sampling strategies, so no single protocol can serve as the reference answer. Second, open-ended experimental plans resist representation as structured labels, which makes objective scoring difficult. To address these challenges, we introduce \textbf{\baby}, a benchmark that reformulates open-ended geochemical experimental design as a \textit{hypothesis-based experiment selection} problem, \ie a structured multi-label classification problem. We abstract concrete instrument names and procedural details into a shared taxonomy of 19 experimental types across five mineral-system processes: \textit{Source, Transport, Deposition, Modification,} and \textit{Preservation}. This taxonomy yields deterministic multi-label targets while accommodating diverse experimental protocols. Constructed from published ore-deposit literature and verified by domain experts, \baby comprises 277 cases and 1,572 reference experimental plans. Given a geological hypothesis and initial context, an LLM agent selects the relevant tasks and generates concrete experimental plans, which are scored by multi-label classification metrics and an LLM-as-a-Judge evidence-coverage metric. We benchmark four open-source agent harnesses, each paired with seven frontier LLMs. The results show that current agents generally achieve high recall but low precision, tending to propose more experiments than are necessary to test the hypothesis.

\vspace{3pt} \noindent \github \quad \textbf{Data and Code}: \url{https://github.com/wangbing1416/MineralExp}

\vspace{3pt} \noindent \web \quad \textbf{Demo}: \url{https://wangbing1416.github.io/projects/mineralexp}
\end{abstract}

\ccsdesc[500]{Applied computing~Earth and atmospheric sciences}
\ccsdesc[500]{Computing methodologies~Natural language processing}

\keywords{large language model, LLM agent, AI for science, mineral exploration, benchmark}


\maketitle

\section{Introduction}
\label{sec:introduction}

Real-world mineral exploration fundamentally relies on formulating and verifying a variety of geological hypotheses about how ore deposits form, evolve, and are preserved \cite{mccuaig2010translating}. Each hypothesis typically breaks down into several distinct, individually testable questions. For example, if an intrusion is proposed to be genetically linked to an ore body, one must ask whether their ages are compatible, and whether they share a common material source, \etc
Typically, resolving such questions requires designing a set of complementary geochemical experiments, \eg fluid-inclusion analysis and mineral trace-element chemistry~\cite{wilkinson2001fluid,cook2016trace,huston2023light,chellemichou2023upb}.
Accordingly, geochemical experimental design sits at the critical juncture between a working geological hypothesis and the empirical evidence needed to confirm or refute it.

Recently, machine learning and \textit{large language model} (LLM) agents are increasingly being integrated into mineral exploration and geochemical workflows. Machine learning methods have substantially advanced mineral prospectivity mapping and the interpretation of geochemical datasets~\citep{bergen2019machine,zuo2020geodata,yang2024artificial,xu2025recent}. Building on this, domain-specific LLMs and multimodal LLM agents now extend automation to geological knowledge retrieval and remote-sensing target generation~\citep{deng2024k2,yu2024mineagent,huang2025peace,zhang2026phreeqcmcq200}.
While these advances excel at extracting patterns from existing observations or executing predefined simulations, a crucial upstream capability remains largely unexplored: \textbf{automated geochemical experimental design}. Given an initial geological hypothesis and limited field observations, intelligent agents are needed that can autonomously decide which geochemical experiments to conduct before committing to costly, time-consuming laboratory work.

However, developing reliable LLM agents for this task requires rigorous evaluation, and there is currently no benchmark to evaluate automated geochemical experimental design. Unlike existing scientific benchmarks that evaluate code execution and data analysis~\cite{chen2025scienceagentbench,majumder2025discoverybench,lou2025aaar10,reilly2026labbench}, 
geochemical experimental design neither has a single standard answer nor yields easily structured labels. Specifically, establishing such a benchmark faces two fundamental challenges. 
First, \textbf{geochemical experiments exhibit high methodological diversity and subjectivity}. Analytical methods do not map one-to-one onto scientific questions: to verify the same hypothesis, different geologists may choose different target minerals or sampling strategies, depending on their expertise. Second, \textbf{experimental design is inherently open-ended and resists representation with structured labels}. A valid experimental plan requires complex reasoning that links the sampled materials and the hypothesis under test. Because multiple concrete protocols can validly address the same hypothesis, comparing a free-text design against a single reference protocol cannot yield a deterministic and objective evaluation metric.

\begin{figure}[t]
    \centering
    \includegraphics[width=0.95\columnwidth]{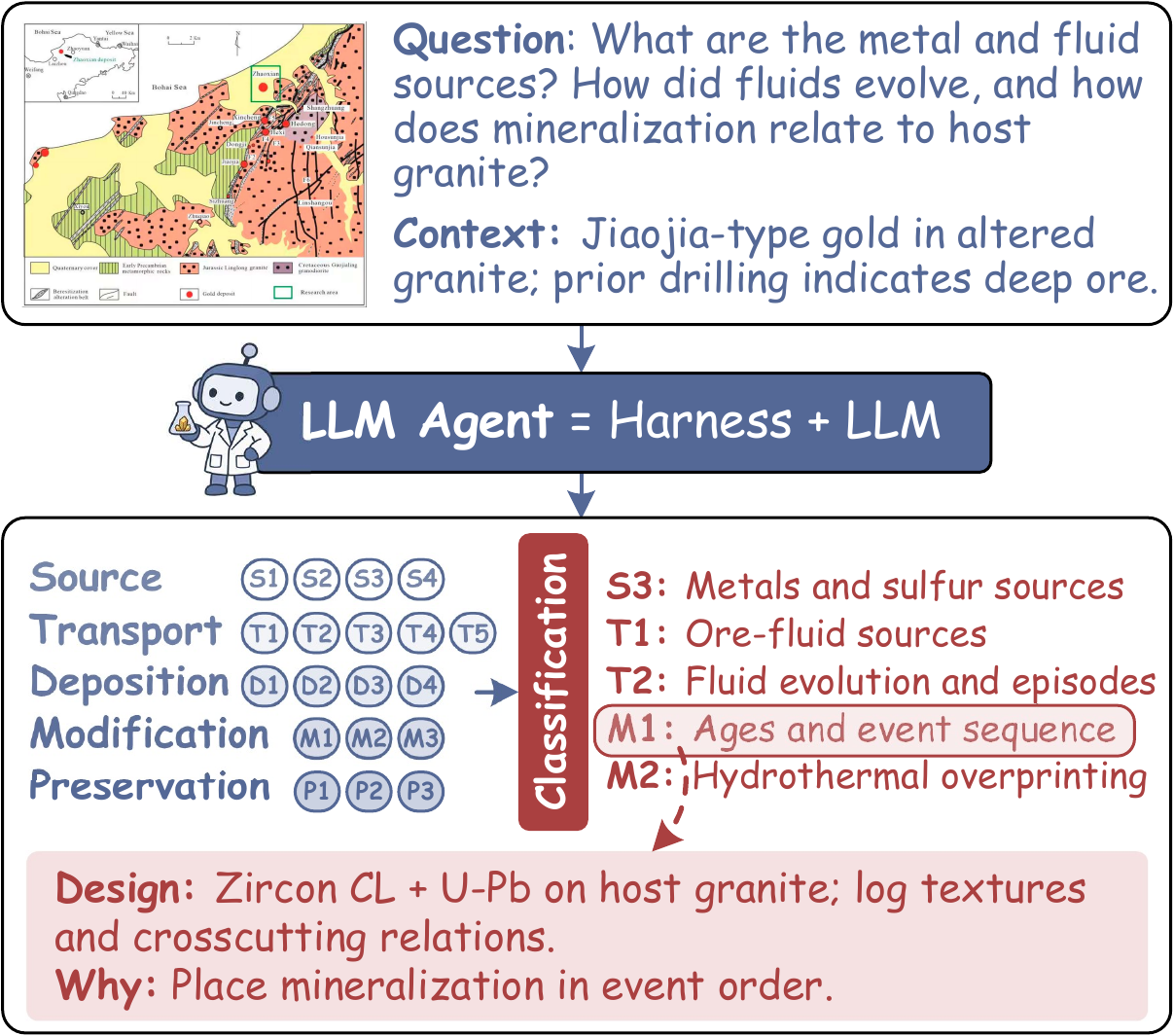}
    \caption{A representative example from \baby. Given a geological hypothesis and its initial context, an LLM agent selects the relevant experiments and proposes the corresponding experimental plans.}
    \label{fig:case}
\end{figure}

To address these challenges, we introduce \textbf{MineralExp}, a benchmark that reformulates open-ended geochemical experimental design as a \textit{hypothesis-based experiment selection} problem, \ie a structured multi-label classification problem. Specifically, to overcome protocol subjectivity and enable structured labeling, we abstract concrete instrument names and procedural details into a shared taxonomy of 19 experimental types, organized across five mineral-system processes: \textit{Source, Transport, Deposition, Modification,} and \textit{Preservation}~\citep{chamberlin1965method,sibson1992earthquake,deng2016gold,zuo2017machine,qaderi2025translation}. The taxonomy captures \textit{what an investigation seeks to establish}, thereby providing deterministic multi-label targets for evaluation while accommodating the diversity of specific experimental designs.
We construct \baby by retrospectively and automatically extracting scientific hypotheses, specific experimental plans, and experimental observations from published ore-deposit literature, followed by format checking and human expert review. This process yields \textbf{277} expert-reviewed cases and \textbf{1,572} reference experimental plans. In each case, an LLM agent receives a structured input comprising a geological hypothesis, the initial geological context, and the available conditions. The agent must then select the relevant subset of the 19 experimental types and generate a structured experimental plan for each selected task, specifying the concrete experimental design, its rationale, and the expected evidence.

Building on this formulation of \baby, we design two categories of evaluation metrics: (1) multi-label classification metrics for experiment selection, \eg recall and F1; and (2) an LLM-as-a-Judge evaluation~\citep{zheng2023judging,jiang2026codejudgebench} of evidence coverage, which measures the proportion of reference experiments covered by the plans an agent generates.
We evaluate four agent harnesses: two general-purpose ones, OpenCode~\citep{anomaly2026opencode} and OpenHands~\citep{wang2025openhands}, and two specialized for experimental design in biology and chemistry, Biomni~\citep{huang2025biomni} and ChemToolAgent~\citep{yu2025tooling}. Each harness is paired with seven LLMs from different model families, \eg GPT-6.1 Sol and Claude Sonnet 5.5. The results show that these state-of-the-art LLMs and agent harnesses generally achieve high recall but low precision, \ie they tend to propose more experiments than are necessary to test the hypothesis. This leaves considerable room for improvement in future LLM agents dedicated to geochemical experimental design.

Our main contributions are summarized as follows:
\begin{itemize}
    \item We formulate geochemical experimental design as a multi-label experiment selection problem spanning 19 experiments across five mineral-system processes, which bridges open-ended protocol design and objective evaluation.
    \item We construct \baby, a benchmark of 277 expert-reviewed, literature-grounded cases with 1,572 reference experimental plans, which preserves the full link between geological hypotheses and the experiments designed to test them.
    \item We provide a standardized evaluation pipeline and benchmark a range of contemporary LLMs and agent harnesses, revealing that current agents struggle to produce precise, context-aware geochemical experimental plans.
\end{itemize}

\section{Related Work}
\label{sec:related_work}

In this section, we review related works on LLM agents for mineral exploration and general scientific discovery.

\subsection{Experiments in Mineral Exploration}

The mineral systems approach organizes mineral exploration around the geological processes that form, modify, and preserve ore deposits, linking deep Earth processes to observable targeting criteria~\cite{mccuaig2010translating}. Testing hypotheses within this framework relies on complementary geochemical experiments, \eg fluid-inclusion microthermometry and mineral trace-element chemistry~\citep{wilkinson2001fluid,cook2016trace,huston2023light,chellemichou2023upb}. Crucially, analytical methods do not map one-to-one onto geological questions. For example, an isotopic signature may be consistent with several candidate sources without singling one out, and a radiometric date must be interpreted in its geological context before it can be assigned to a mineralization event. These characteristics motivate the design of \baby, which abstracts concrete instrument names into 19 purpose-driven experimental types across five mineral-system processes.

\subsection{LLM Agents for Mineral Exploration}

Machine learning has been widely applied to downstream geoscientific inference over already-collected data, \eg mineral prospectivity mapping and geochemical anomaly interpretation~\cite{bergen2019machine,zuo2020geodata,sun2022machine,yang2024artificial,petrelli2024machine,xu2025recent}, as well as sequential field-measurement acquisition~\cite{mern2023the}. More recently, LLMs agents have extended automation to ontology-guided knowledge extraction from mineral-exploration and geochemical literature~\cite{zhang2026ontogrc,jahin2026articleminer}, geoscientific question answering and primary-record reasoning~\cite{deng2024k2,eigenform2026groundtruth}, multimodal interpretation of geological maps and remote-sensing imagery~\cite{yu2024mineagent,yu2025stacot,huang2025peace}, and aqueous-geochemistry simulation~\cite{zhang2026phreeqcmcq200}. For example, MineAgent~\cite{yu2024mineagent} leverages multimodal LLMs to jointly reason over geological maps and hyperspectral imagery for mineral exploration targeting. However, these efforts primarily interpret available observations, extract static text, or run predefined simulators. In contrast, \baby targets the upstream planning decision that precedes laboratory execution: autonomously deciding which geochemical experiments to conduct in order to test a working geological hypothesis.

\subsection{General Scientific Agents}

Beyond geoscience, reasoning and tool-using LLM agents have shown strong capabilities in chemical synthesis and research workflows~\cite{boiko2023autonomous,bran2024augmenting,wang2026on}, which has motivated benchmarks for scientific coding, data-driven discovery, and interactive simulation environments~\cite{jansen2024discoveryworld,chen2025scienceagentbench,majumder2025discoverybench}. Closer to our work, recent benchmarks evaluate experimental planning in biological protocol reconstruction~\cite{odonoghue2023bioplanner}, wet-lab next-experiment ranking~\cite{reilly2026labbench}, and general research design~\cite{lou2025aaar10,abramovich2025ablationbench,liu2026can}. For example, LabBench~\cite{reilly2026labbench} evaluates agents on wet-lab experimental reasoning by asking them to rank candidate next experiments against recorded laboratory decisions. While these benchmarks establish experimental planning as a critical capability in computational or standardized wet-lab settings, geochemical investigations lack fixed procedural recipes and readily structured labels. \baby bridges this gap by combining a multi-label taxonomy of mineral tasks with LLM-as-a-Judge evidence coverage, grounded in expert-reviewed plans drawn from the literature.

\section{Task Formulation and Taxonomy}
\label{sec:task}

In this section, we present the geoscientific rationale behind automated geochemical experimental design, formally define the task in \baby, and detail the five-process, 19-task mineral taxonomy.

\subsection{Scientific Scope and Reasoning Challenges}

Testing ore-forming mechanisms requires translating competing genetic hypotheses into testable propositions that independent geological evidence can distinguish. In real-world mineral exploration, similar mineralization patterns, mineral assemblages, and geochemical signatures may arise from distinct geological processes, including metal sourcing, fluid generation and transport, metal precipitation, post-ore modification, and deposit preservation~\citep{chamberlin1965method,sibson1992earthquake,deng2016gold,zuo2017machine,qaderi2025translation}. Experimental planning must therefore determine which mechanisms need empirical testing, which diagnostic observables can discriminate among the competing alternatives, and which specific measurements are required.

Designing such experiments is further complicated by the heterogeneity of ore-forming records, which are often overprinted by successive geological events. Geochronological, isotopic, mineral-scale, fluid-inclusion, and structural datasets differ in temporal resolution, spatial scale, and process specificity, so their valid interpretation must be constrained by mineral generations, reaction conditions, and spatiotemporal relationships~\citep{eidsvik2013value,engi2017significant}. For example, to determine whether invisible gold in sulfides occurs as lattice-bound solid solution or as nanoparticulate inclusions, one must first assess whether the spatial resolution of existing LA-ICP-MS trace-element maps is sufficient, and if not, whether nanometer-scale transmission electron microscopy is warranted. Automated geochemical experimental design therefore targets \textit{empirical distinguishability}: it matches mechanistic ambiguities to appropriate sample materials, analytical methods, and verification criteria, rather than merely reciting a fixed analytical recipe.

\subsection{Problem Formulation}

To make open-ended geochemical experimental design amenable to rigorous benchmarking while preserving its scientific depth, \baby formulates each case as a joint \textit{hypothesis-conditioned experiment selection} and \textit{plan generation} problem.

\vspace{2pt} \noindent
\textbf{Input specification.}
Each benchmark instance provides a structured input tuple $\mathcal{X} = (\mathbf{H}, \mathbf{C}, \mathcal{A})$, derived from published ore-deposit literature~\citep{groves1998orogenic,goldfarb2015orogenic,groves2018structural,deng2020situ,qiu2020giant,qiu2023low}:
\begin{itemize}
    \item \textbf{Geological hypothesis $\mathbf{H}$:} The core metallogenic proposition, or the competing genetic mechanisms, under investigation, \eg magmatic-hydrothermal versus metamorphic sources of fluid and metals.
    \item \textbf{Initial geological context $\mathbf{C}$:} Prior field and petrographic observations that establish the baseline geological setting, including host-rock lithology, structural controls, alteration zoning, and mineral paragenesis.
    \item \textbf{Available conditions $\mathcal{A}$:} The geological materials and constraints available before laboratory testing, \eg drill cores and specific mineral separates.
\end{itemize}

\vspace{2pt} \noindent
\textbf{Taxonomy and agent output.}
Let $\mathcal{T} = \{t_1, t_2, \dots, t_{19}\}$ denote our shared taxonomy of 19 experimental types, organized across five mineral-system processes: \textit{Source, Transport, Deposition, Modification,} and \textit{Preservation}. Given the input $\mathcal{X}$, an LLM agent should produce a two-level output $(\widehat{\mathcal{Y}}, \widehat{\mathcal{P}})$:
\begin{itemize}
    \item \textbf{Multi-label experiment selection $\widehat{\mathcal{Y}} \subseteq \mathcal{T}$:} The subset of experiments necessary to test the hypothesis $\mathbf{H}$ under context $\mathbf{C}$ and conditions $\mathcal{A}$. Each selected experiment $t \in \widehat{\mathcal{Y}}$ represents a discrete experimental objective, \ie what the investigation seeks to establish.
    \item \textbf{Structured experimental plans $\widehat{\mathcal{P}} = \{\mathbf{p}_t\}_{t \in \widehat{\mathcal{Y}}}$:} For each selected experiment $t \in \widehat{\mathcal{Y}}$, the agent generates a concrete experimental plan $\mathbf{p}_t = (\mathbf{d}_t, \mathbf{r}_t, \mathbf{e}_t)$ that specifies (1) the \textit{experimental design} $\mathbf{d}_t$, \eg the target sample or mineral generation and the analytical technique; (2) the \textit{scientific rationale} $\mathbf{r}_t$, which addresses the mechanistic ambiguity; and (3) the \textit{expected evidence} $\mathbf{e}_t$, together with the conditional interpretations that would confirm or refute the hypothesis.
\end{itemize}

\vspace{2pt} \noindent
\textbf{Evaluation target.}
For each input $\mathcal{X}$, agent outputs are evaluated against an expert-reviewed reference task set $\mathcal{Y}^* \subseteq \mathcal{T}$ and reference experimental plans $\mathcal{P}^*$ extracted from the source literature. Task-level selection quality is measured by multi-label classification metrics between $\widehat{\mathcal{Y}}$ and $\mathcal{Y}^*$, \eg recall and F1. Protocol-level fidelity is assessed by LLM-as-a-Judge evidence coverage, defined as the proportion of reference plans in $\mathcal{P}^*$ that are covered by $\widehat{\mathcal{P}}$.

\subsection{Specific Experimental Taxonomy}

To provide a multi-label structure, \baby organizes its 19 experimental types around the five-stage lifecycle of an ore deposit, as detailed in Tables~\ref{tab:processes} and~\ref{tab:taxonomy}. The five stages are described below.

\vspace{2pt} \noindent
\textbf{Source (S1-S4):} \textit{where did the ore-forming materials come from?}
Source tasks identify the origin of magmas, fluids, and metals (S1, S3, S4), and assess whether a candidate source rock was fertile enough to supply the deposit (S2). Because different geological reservoirs can share similar chemical traits, experiments in this category typically rely on isotopic fingerprinting, \eg S, Pb, or Sr--Nd isotopes, and whole-rock geochemistry, to distinguish competing sources.

\vspace{2pt} \noindent
\textbf{Transport (T1-T5):} \textit{how did fluids carry metals from the source to the deposit?}
Transport tasks investigate the origins and evolution of the carrier fluids (T1, T2), their physical and chemical conditions, \eg temperature and pressure (T3), the mechanisms by which metals were dissolved and carried (T4), and the fault or fracture pathways through which the fluids migrated (T5). Typical experiments analyze microscopic fluid inclusions trapped in crystals, together with mineral chemistry, to reconstruct fluid migration.

\vspace{2pt} \noindent
\textbf{Deposition (D1-D4):} \textit{how and why did metals precipitate to form an orebody?}
Deposition tasks examine the state in which metals are hosted within minerals (D1), the chemical reactions that triggered metal precipitation (D2), the efficiency of element enrichment (D3), and the spatial zoning patterns across the deposit (D4). Typical experiments use high-resolution micro-analytical mapping, \eg laser ablation or electron microprobe analysis, to reveal how fluid cooling or fluid-rock reactions concentrated the metals.

\vspace{2pt} \noindent
\textbf{Modification (M1-M3):} \textit{when did the deposit form, and was it altered later?}
Modification tasks establish the absolute age and sequence of mineralization events (M1), and test whether later heating or fluid pulses overprinted the earlier record (M2) or remobilized existing metals (M3). Because later geological events can reset isotopic clocks, experiments here combine microtextural screening with radiometric dating, \eg U-Pb or Re-Os geochronology, on specific mineral generations.

\vspace{2pt} \noindent
\textbf{Preservation (P1-P3):} \textit{has the orebody survived erosion and weathering?}
Preservation tasks assess the post-mineralization cooling and uplift history (P1), near-surface weathering and secondary enrichment (P2), and the likelihood that concealed orebodies remain intact at depth (P3). Typical experiments include low-temperature thermochronology and depth-profile geochemical analysis.

\section{The \baby Benchmark}
\label{sec:benchmark}

Building on the task formulation in Sec.~\ref{sec:task}, this section details the construction pipeline of \baby, analyzes its composition and task distribution, and describes the evaluation protocol.

\subsection{Dataset Construction Pipeline}
\label{sec:construction}

\begin{figure}[t]
    \centering
    \includegraphics[width=0.95\columnwidth]{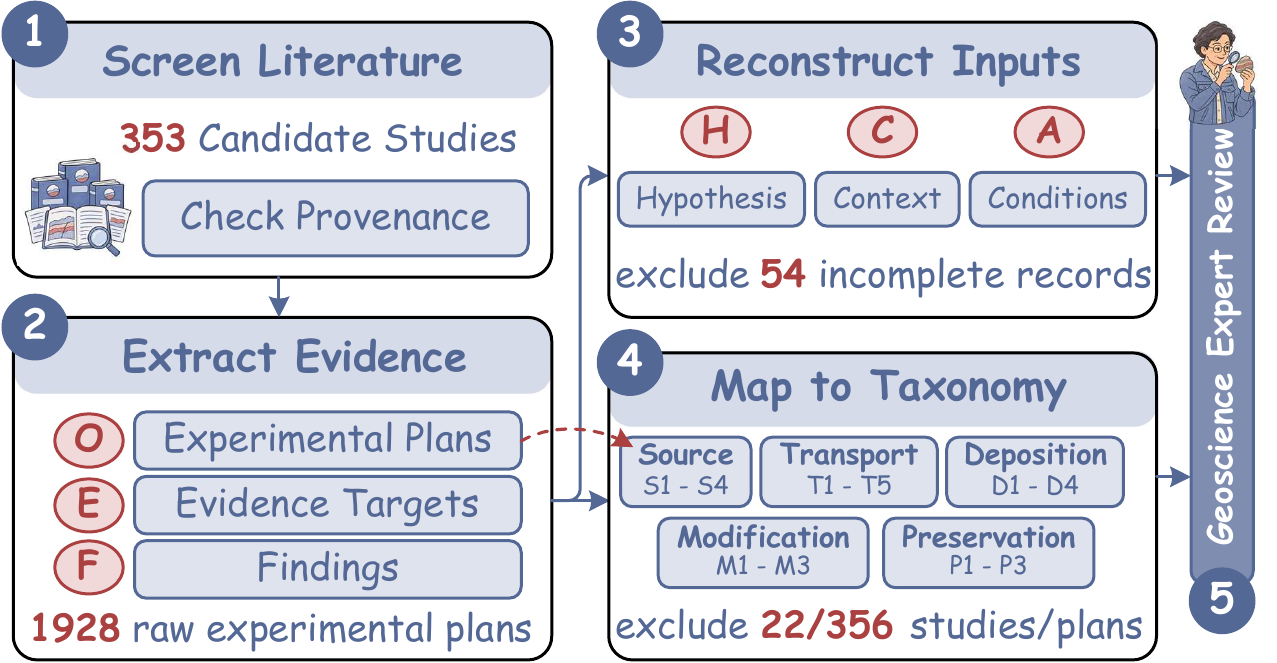}
    \caption{Overall construction pipeline of \baby. From 353 candidate studies, we extract $(O, E, F)$ records to separately construct decontaminated inputs $\mathcal{X}_i = (\mathbf{H}_i, \mathbf{C}_i, \mathcal{A}_i)$ and mapped reference targets $(\mathcal{O}_i^*, \mathcal{Y}_i^*)$, followed by expert review and validation to yield 277 cases and 1,572 reference plans.}
    \label{fig:framework}
\end{figure}

We construct \baby through a rigorous five-stage pipeline that converts published ore-deposit studies into retrospective experimental planning cases. Throughout the pipeline, LLMs are used strictly to draft structured records from source texts, which are then verified by human experts. For clarity, the overall pipeline of \baby is depicted in Fig.~\ref{fig:framework}.

\vspace{2pt} \noindent
\textbf{Literature collection and source screening.}
Candidate studies are curated from peer-reviewed international journals in geochemistry and mineral exploration, supplemented by public dissertations and geological survey reports. Each document undergoes provenance verification to confirm its bibliographic identity and to identify its primary research contributions. To respect copyright constraints, \baby releases only reconstructed task inputs and structured annotations, rather than the raw full-text articles.

\vspace{2pt} \noindent
\textbf{Extraction of experimental plans, evidence targets, and findings.}
For each study, we extract a triplet curation record $(\mathcal{P}^*, \mathcal{E}, \mathcal{F})$, with every component anchored to locators in the source text:
\begin{itemize}
    \item \textbf{Plans $\mathcal{P}^*$:} The concrete actions reported by the authors, such as sampling, sample preparation, petrographic characterization, and geochemical measurement.
    \item \textbf{Evidence targets $\mathcal{E}$:} The specific mechanistic ambiguities or questions that the actions in $\mathcal{P}$ are intended to resolve.
    \item \textbf{Findings $\mathcal{F}$:} The resulting empirical measurements, analytical uncertainties, and author interpretations in the study.
\end{itemize}

\vspace{2pt} \noindent
\textbf{Retrospective input reconstruction and decontamination.}
From each curated record, we reconstruct the input tuple $\mathcal{X}_i = (\mathbf{H}_i, \mathbf{C}_i, \mathcal{A}_i)$ defined in Sec.~\ref{sec:task}. The hypothesis $\mathbf{H}_i$ captures the central metallogenic question of the study; the initial context $\mathbf{C}_i$ summarizes the prior geological setting; and the available conditions $\mathcal{A}_i$ specify the starting sample materials and field constraints, \eg accessible ore and host-rock specimens, drill cores, or field exposures. To prevent target leakage, we strictly exclude the newly obtained experimental findings $\mathcal{F}$ and the conclusions of the study from $\mathcal{X}_i$. A candidate case is retained only if all three input fields are non-empty and explicitly supported by the source text.

\vspace{2pt} \noindent
\textbf{Purpose-driven mapping to the task taxonomy.}
For each case $i$, we filter out preparatory or out-of-scope plans that do not address any of the 19 tasks in $\mathcal{T}$, and the remaining plans form the scored set of reference experimental plans $\mathcal{P}_i^*$. Each reference plan $p \in \mathcal{P}_i^*$ is then mapped to its corresponding task annotations $\mathbf{m}(p) \subseteq \mathcal{T}$ according to its sampled materials and evidence targets $\mathcal{E}$. For example, zircon U--Pb dating aimed at constraining the depositional age of the host strata is distinguished from dating mineralization events, whereas routine petrography or cathodoluminescence imaging is treated as a preparatory step. The case-level reference task set $\mathcal{Y}_i^* \subseteq \mathcal{T}$ is then obtained as the union over all retained plans:
\begin{equation}
    \mathcal{Y}_i^* = \bigcup_{p \in \mathcal{P}_i^*} \mathbf{m}(p).
    \label{eq:targetset}
\end{equation}
Under this formulation, multiple analytical plans pursued for the same evidence target contribute only a single positive task label to $\mathcal{Y}_i^*$, while a multipurpose plan may map to more than one task when warranted.

\vspace{2pt} \noindent
\textbf{Expert review and quality validation.}
Finally, domain experts in economic geology and geochemistry independently reviewed all candidate cases, the reconstructed inputs $\mathcal{X}_i$, and the plan-to-task mappings $(\mathcal{P}_i^*, \mathcal{Y}_i^*)$. Through expert adjudication, we refined the boundary rules and excluded studies outside the task scope. For example, constraints on the depositional age of host strata are admitted under task M1 when they are used to bracket the timing of mineralization. Automated schema validation was then applied to verify input completeness, label consistency between $\mathcal{P}_i^*$ and $\mathcal{Y}_i^*$, and traceability to source hashes.

Starting from an initial archive of 353 candidate studies, the pipeline removed 54 records with incomplete or unsupported input fields and 22 records outside the target task scope, yielding \textbf{277 expert-reviewed cases}. Of the 1,928 plans extracted from these cases, \textbf{1,572} are retained as scored reference plans $\mathcal{P}_i^*$.

\begin{table}[t]
    \centering
    \small
    \setlength{\tabcolsep}{5pt}
    \renewcommand\arraystretch{0.9}
    \caption{Process-level composition of \baby. }
    \label{tab:processes}
    \begin{tabular}{m{1.5cm}<{\centering}m{1.3cm}<{\centering}m{0.9cm}<{\centering}m{1.8cm}<{\centering}}
        \toprule
        Process & Task IDs & Tasks & Positive Cases \\
        \midrule
        Source       & S1--S4 & 4 & 160 \\
        Transport    & T1--T5 & 5 & 170 \\
        Deposition   & D1--D4 & 4 & 146 \\
        Modification & M1--M3 & 3 & 143 \\
        Preservation & P1--P3 & 3 & 46 \\
        \midrule
        Total & - & 19 & 277 \\
        \bottomrule
    \end{tabular}
\end{table}

\subsection{Dataset Composition and Statistics}
\label{sec:datasetcomposition}

The 277 cases in \baby comprise 1,572 reference experimental plans and 968 positive case-task assignments. On average, each case involves 3.49 experiments, reflecting the multi-faceted nature of real-world hypothesis testing.
Tables~\ref{tab:processes} and~\ref{tab:taxonomy} summarize the distribution of cases across the five mineral-system processes and the 19 experimental types. The categories capture distinct scientific objectives: for example, isotope analyses that trace metal provenance (S3) are distinguished from those that constrain fluid sources (T1), and mineralization dating (M1) is separated from the recognition of multistage hydrothermal overprinting (M2).
Reflecting the natural research priorities of the ore-deposit literature, the label distribution is long-tailed. Core tasks such as metal/sulfur provenance (S3, 128 cases), mineralization dating (M1, 108 cases), and fluid $P$--$T$--volatile characterization (T3, 99 cases) occur frequently, whereas specialized tasks such as metamorphic metal release (S4) and gold remobilization (M3) are rare in the current release. We nevertheless retain all 19 tasks in the shared prediction space $\mathcal{T}$, which tests whether agents can avoid over-selecting rare experiments that the geological context does not support.

\begin{table}[t]
    \centering
    \small
    \renewcommand\arraystretch{0.9}
    \caption{The 19 experimental types in \baby and their corresponding number of positive cases.}
    \label{tab:taxonomy}
    \begin{tabular}{m{0.4cm}<{\centering}m{6.3cm}<{\centering}m{0.5cm}<{\centering}}
        \toprule
        ID & Experimental Types & Cases \\
        \midrule
        S1 & Magma sources and evolution & 50 \\
        S2 & Magmatic metallogenic potential & 19 \\
        S3 & Sources of metals and sulfur & 128 \\
        S4 & Metamorphic fluids and metal release & 1 \\
        \midrule
        T1 & Ore-fluid sources & 95 \\
        T2 & Fluid evolution and episodes & 95 \\
        T3 & Fluid pressure, temperature, and volatiles & 99 \\
        T4 & Metal loading and transport & 11 \\
        T5 & Structurally controlled fluid pathways & 57 \\
        \midrule
        D1 & Gold occurrence mechanisms & 64 \\
        D2 & Gold precipitation reactions & 23 \\
        D3 & Element transfer and enrichment efficiency & 21 \\
        D4 & Mineralization zoning and spatial response & 94 \\
        \midrule
        M1 & Mineralization ages and event sequence & 108 \\
        M2 & Multistage hydrothermal overprinting & 55 \\
        M3 & Gold remobilization and reprecipitation & 1 \\
        \midrule
        P1 & Post-ore cooling and exhumation & 1 \\
        P2 & Supergene modification and secondary enrichment & 6 \\
        P3 & Preservation and prediction of concealed mineralization & 41 \\
        \bottomrule
    \end{tabular}
\end{table}

\subsection{Structured Agent Response Schema}
\label{sec:structuredresponses}

To enforce explicit reasoning when producing $(\widehat{\mathcal{Y}}_i, \widehat{\mathcal{P}}_i)$, our evaluation requires agents to return a structured response containing the selected task set $\widehat{\mathcal{Y}}_i \subseteq \mathcal{T}$ and 19 task records, one for each task in $\mathcal{T}$. Each record specifies a Boolean selection flag, a decision rationale, and a list of proposed experiments. For each unselected task $t \notin \widehat{\mathcal{Y}}_i$, the agent must provide an exclusion rationale and leave the experiment list empty. Conversely, each selected task $t \in \widehat{\mathcal{Y}}_i$ must include at least one concrete experimental plan $\mathbf{p}_t = (\mathbf{d}_t, \mathbf{r}_t, \mathbf{e}_t) \in \widehat{\mathcal{P}}_i$, which is instantiated through seven mandatory fields. These include the \textit{experiment name}; the \textit{concrete experimental design} $\mathbf{d}_t$, covering target materials, analytical methods, procedures, and quality controls; the \textit{scientific rationale} $\mathbf{r}_t$, explaining why the experiment is needed; and the \textit{expected evidence and interpretations} $\mathbf{e}_t$, covering expected observables, conditions that would support the hypothesis, conditions that would weaken it, and analytical limitations.

\subsection{Evaluation Metrics}
\label{sec:evaluationinterface}

Following Sec.~\ref{sec:task}, we evaluate the agent outputs $(\widehat{\mathcal{Y}}_i, \widehat{\mathcal{P}}_i)$ against the ground-truth references $(\mathcal{Y}_i^*, \mathcal{P}_i^*)$ using two groups of metrics.

\vspace{2pt} \noindent
\textbf{Multi-label experiment selection metrics.}
To compare the predicted experiment set $\widehat{\mathcal{Y}}_i$ with the expert reference $\mathcal{Y}_i^*$ across all $N=277$ cases, we compute micro-precision $P$, micro-recall $R$, micro-F1 $F_{1}$, macro-F1, and exact-set accuracy. The micro-averaged metrics are defined as
\begin{equation}
    P = \frac{\mathrm{TP}}{\mathrm{TP}+\mathrm{FP}}, \quad
    R = \frac{\mathrm{TP}}{\mathrm{TP}+\mathrm{FN}}, \quad
    F_{1} = \frac{2\mathrm{TP}}{2\mathrm{TP}+\mathrm{FP}+\mathrm{FN}},
\end{equation}
where $\mathrm{TP}$, $\mathrm{FP}$, and $\mathrm{FN}$ are aggregated over all case--task pairs. Exact-set accuracy is the fraction of cases for which $\widehat{\mathcal{Y}}_i = \mathcal{Y}_i^*$, and macro-F1 is the unweighted mean of the per-task F1 scores over $\mathcal{T}$. We also compute process-wise metrics for each of the five mineral-system processes by projecting the label space onto that process's tasks while retaining all 277 cases, so that false-positive selections in processes with no reference tasks are properly penalized.

\vspace{2pt} \noindent
\textbf{LLM-as-a-Judge evidence coverage.}
To assess whether the agent's experimental plans $\widehat{\mathcal{P}}_i$ adequately capture the empirical investigations conducted in the source study, we adopt an LLM-as-a-Judge protocol~\citep{zheng2023judging,jiang2026codejudgebench}. For each case $i$, the judge compares the proposed plans $\widehat{\mathcal{P}}_i$ against the reference plans $\mathcal{P}_i^*$. A reference plan $p \in \mathcal{P}_i^*$ is deemed covered, i.e., $\mathbb{I}_{\mathrm{cov}}(p, \widehat{\mathcal{P}}_i) = 1$, if the agent proposes a valid experimental design that targets a compatible geological material and analytical objective and is capable of yielding equivalent diagnostic evidence. The evidence coverage is formulated as
\begin{equation}
    \mathrm{Cov} = \frac{1}{\sum \nolimits_{i=1}^{N} |\mathcal{P}_i^*|} \sum \nolimits_{i=1}^{N} \sum \nolimits_{p \in \mathcal{P}_i^*} \mathbb{I}_{\mathrm{cov}}\big(p, \widehat{\mathcal{P}}_i\big).
\end{equation}

\section{Experimental Evaluation}
\label{sec:experiments}

In this section, we conduct extensive experiments to evaluate how well contemporary LLM agents perform on \baby. 

\subsection{Experimental Settings}
\label{subsec:exp_settings}

\noindent
\textbf{Agent harnesses.} 
We evaluate four representative open-source agent harnesses on \baby, which fall into two categories: two general-purpose task-execution harnesses, \ie OpenCode~\cite{anomaly2026opencode} and OpenHands~\cite{wang2025openhands}, and two domain-specific scientific harnesses designed for experimental design in biology and chemistry, \ie Biomni~\cite{huang2025biomni} and ChemToolAgent~\cite{yu2025tooling}. Detailed descriptions of these harnesses are provided in Appendix~\ref{app:harnesses}.

\vspace{2pt} \noindent
\textbf{Foundation LLMs.} 
We pair each agent harness with seven frontier LLMs spanning different model families and parameter scales: GPT-6.1 Sol, GPT-6 Luna, Claude Sonnet 5.5, Claude Haiku 5.5, DeepSeek-V4.1-Flash, Qwen3.8-27B, and Sage-A1.5. Their detailed model cards are provided in Appendix~\ref{app:llms}.

\vspace{2pt} \noindent
\textbf{Evaluation protocol.} 
We evaluate all $4 \times 7 = 28$ agent configurations on \baby. For each case, the agent receives the decontaminated input tuple $\mathcal{X}_i = (\mathbf{H}_i, \mathbf{C}_i, \mathcal{A}_i)$ and generates both the selected experimental tasks $\widehat{\mathcal{Y}}_i$ and the corresponding experimental plans $\widehat{\mathcal{P}}_i$. Implementation details of the evaluation pipeline and metrics are provided in Appendices~\ref{app:impl_details} and~\ref{app:metrics}, respectively.

\begin{table*}[t]
\centering
\renewcommand\arraystretch{0.98}
\caption{Experimental results of 4 open-source agent harnesses and 7 foundation LLMs on our benchmark \baby.}
\label{tab:overall}
\small
\setlength{\tabcolsep}{3pt}
\begin{tabular}{m{0.5cm}<{\centering}m{2.8cm}<{\centering}m{0.9cm}<{\centering}m{0.9cm}<{\centering}m{0.9cm}<{\centering}m{0.9cm}<{\centering}m{0.9cm}<{\centering}m{0.9cm}<{\centering}m{0.9cm}<{\centering}m{0.9cm}<{\centering}m{0.9cm}<{\centering}m{0.9cm}<{\centering}m{0.9cm}<{\centering}m{0.9cm}<{\centering}}
\toprule
\multirow{2}{*}{Agent} & \multirow{2}{*}{LLM} & \multicolumn{2}{c}{Source} & \multicolumn{2}{c}{Transport} & \multicolumn{2}{c}{Deposition} & \multicolumn{2}{c}{Modification} & \multicolumn{2}{c}{Preservation} & \multicolumn{2}{c}{\textbf{Overall}} \\
\cmidrule(lr){3-4} \cmidrule(lr){5-6} \cmidrule(lr){7-8} \cmidrule(lr){9-10} \cmidrule(lr){11-12} \cmidrule(lr){13-14}
& & R. & F1 & R. & F1 & R. & F1 & R. & F1 & R. & F1 & R. & F1 \\
\midrule
\multirow{7}{*}{\rotatebox{90}{\textbf{OpenCode}}}
 & GPT-6.1 Sol & \cellcolor{mydarkblue!2}72.08 & \cellcolor{mydarkblue!20}68.77 & \cellcolor{mydarkblue!0}76.19 & \cellcolor{mydarkblue!20}67.16 & \cellcolor{mydarkblue!0}63.86 & \cellcolor{mydarkblue!20}56.09 & \cellcolor{mydarkblue!0}62.80 & \cellcolor{mydarkblue!20}57.87 & \cellcolor{mydarkblue!7}81.25 & \cellcolor{mydarkblue!20}72.90 & \cellcolor{mydarkblue!0}\textbf{70.76} & \cellcolor{mydarkblue!20}\textbf{63.84} \\
 & GPT-6 Luna & \cellcolor{mydarkblue!11}79.19 & \cellcolor{mydarkblue!5}53.70 & \cellcolor{mydarkblue!12}84.03 & \cellcolor{mydarkblue!4}54.01 & \cellcolor{mydarkblue!17}79.21 & \cellcolor{mydarkblue!7}46.51 & \cellcolor{mydarkblue!11}76.83 & \cellcolor{mydarkblue!1}46.24 & \cellcolor{mydarkblue!0}79.17 & \cellcolor{mydarkblue!3}46.91 & \cellcolor{mydarkblue!13}\textbf{80.58} & \cellcolor{mydarkblue!4}\textbf{50.53} \\
 & Claude Sonnet 5.5 & \cellcolor{mydarkblue!10}78.68 & \cellcolor{mydarkblue!20}68.28 & \cellcolor{mydarkblue!11}82.91 & \cellcolor{mydarkblue!19}65.92 & \cellcolor{mydarkblue!8}71.29 & \cellcolor{mydarkblue!13}50.70 & \cellcolor{mydarkblue!7}71.95 & \cellcolor{mydarkblue!10}51.75 & \cellcolor{mydarkblue!20}85.42 & \cellcolor{mydarkblue!17}68.33 & \cellcolor{mydarkblue!9}\textbf{77.89} & \cellcolor{mydarkblue!16}\textbf{60.42} \\
 & Claude Haiku 5.5 & \cellcolor{mydarkblue!17}84.26 & \cellcolor{mydarkblue!6}54.16 & \cellcolor{mydarkblue!11}82.91 & \cellcolor{mydarkblue!11}59.86 & \cellcolor{mydarkblue!4}67.82 & \cellcolor{mydarkblue!12}50.37 & \cellcolor{mydarkblue!11}76.83 & \cellcolor{mydarkblue!2}46.75 & \cellcolor{mydarkblue!13}83.33 & \cellcolor{mydarkblue!6}51.61 & \cellcolor{mydarkblue!11}\textbf{79.03} & \cellcolor{mydarkblue!8}\textbf{53.87} \\
 & DeepSeek-V4.1-Flash & \cellcolor{mydarkblue!20}86.29 & \cellcolor{mydarkblue!0}48.57 & \cellcolor{mydarkblue!20}88.80 & \cellcolor{mydarkblue!0}50.28 & \cellcolor{mydarkblue!20}82.18 & \cellcolor{mydarkblue!0}41.60 & \cellcolor{mydarkblue!20}88.41 & \cellcolor{mydarkblue!0}45.74 & \cellcolor{mydarkblue!0}79.17 & \cellcolor{mydarkblue!0}41.76 & \cellcolor{mydarkblue!20}\textbf{86.36} & \cellcolor{mydarkblue!0}\textbf{46.77} \\
 & Qwen3.8-27B & \cellcolor{mydarkblue!0}70.56 & \cellcolor{mydarkblue!8}56.28 & \cellcolor{mydarkblue!15}85.43 & \cellcolor{mydarkblue!14}62.18 & \cellcolor{mydarkblue!9}72.28 & \cellcolor{mydarkblue!9}48.26 & \cellcolor{mydarkblue!8}73.17 & \cellcolor{mydarkblue!7}50.21 & \cellcolor{mydarkblue!0}79.17 & \cellcolor{mydarkblue!13}61.29 & \cellcolor{mydarkblue!8}\textbf{77.27} & \cellcolor{mydarkblue!11}\textbf{55.78} \\
 & Sage-A1.5 & \cellcolor{mydarkblue!14}81.73 & \cellcolor{mydarkblue!3}51.27 & \cellcolor{mydarkblue!12}83.75 & \cellcolor{mydarkblue!5}54.56 & \cellcolor{mydarkblue!16}78.71 & \cellcolor{mydarkblue!4}44.23 & \cellcolor{mydarkblue!12}78.66 & \cellcolor{mydarkblue!3}47.34 & \cellcolor{mydarkblue!20}85.42 & \cellcolor{mydarkblue!7}53.25 & \cellcolor{mydarkblue!14}\textbf{81.51} & \cellcolor{mydarkblue!4}\textbf{50.22} \\

\hline
\multirow{7}{*}{\rotatebox{90}{\textbf{OpenHands}}}
 & GPT-6.1 Sol & \cellcolor{mydarkblue!0}74.62 & \cellcolor{mydarkblue!18}64.47 & \cellcolor{mydarkblue!6}82.35 & \cellcolor{mydarkblue!19}65.04 & \cellcolor{mydarkblue!9}67.82 & \cellcolor{mydarkblue!18}52.29 & \cellcolor{mydarkblue!5}70.12 & \cellcolor{mydarkblue!20}55.16 & \cellcolor{mydarkblue!7}83.33 & \cellcolor{mydarkblue!20}70.80 & \cellcolor{mydarkblue!3}\textbf{75.72} & \cellcolor{mydarkblue!20}\textbf{60.73} \\
 & GPT-6 Luna & \cellcolor{mydarkblue!16}84.26 & \cellcolor{mydarkblue!1}51.31 & \cellcolor{mydarkblue!16}89.36 & \cellcolor{mydarkblue!3}52.90 & \cellcolor{mydarkblue!18}82.18 & \cellcolor{mydarkblue!5}44.21 & \cellcolor{mydarkblue!18}85.98 & \cellcolor{mydarkblue!5}46.46 & \cellcolor{mydarkblue!20}87.50 & \cellcolor{mydarkblue!4}50.00 & \cellcolor{mydarkblue!17}\textbf{86.16} & \cellcolor{mydarkblue!4}\textbf{49.36} \\
 & Claude Sonnet 5.5 & \cellcolor{mydarkblue!0}76.14 & \cellcolor{mydarkblue!20}66.96 & \cellcolor{mydarkblue!7}81.79 & \cellcolor{mydarkblue!20}65.11 & \cellcolor{mydarkblue!12}71.78 & \cellcolor{mydarkblue!20}53.60 & \cellcolor{mydarkblue!0}64.02 & \cellcolor{mydarkblue!11}50.48 & \cellcolor{mydarkblue!0}81.25 & \cellcolor{mydarkblue!18}68.42 & \cellcolor{mydarkblue!3}\textbf{75.52} & \cellcolor{mydarkblue!19}\textbf{60.51} \\
 & Claude Haiku 5.5 & \cellcolor{mydarkblue!15}83.76 & \cellcolor{mydarkblue!9}57.79 & \cellcolor{mydarkblue!0}77.87 & \cellcolor{mydarkblue!14}61.23 & \cellcolor{mydarkblue!0}53.96 & \cellcolor{mydarkblue!14}49.66 & \cellcolor{mydarkblue!7}72.56 & \cellcolor{mydarkblue!12}50.32 & \cellcolor{mydarkblue!7}83.33 & \cellcolor{mydarkblue!20}71.43 & \cellcolor{mydarkblue!0}\textbf{73.45} & \cellcolor{mydarkblue!14}\textbf{56.81} \\
 & DeepSeek-V4.1-Flash & \cellcolor{mydarkblue!20}86.80 & \cellcolor{mydarkblue!0}50.74 & \cellcolor{mydarkblue!20}92.16 & \cellcolor{mydarkblue!0}50.62 & \cellcolor{mydarkblue!20}85.64 & \cellcolor{mydarkblue!0}40.95 & \cellcolor{mydarkblue!19}87.80 & \cellcolor{mydarkblue!0}43.31 & \cellcolor{mydarkblue!13}85.42 & \cellcolor{mydarkblue!0}44.32 & \cellcolor{mydarkblue!20}\textbf{88.64} & \cellcolor{mydarkblue!0}\textbf{46.77} \\
 & Qwen3.8-27B & \cellcolor{mydarkblue!0}74.62 & \cellcolor{mydarkblue!6}55.16 & \cellcolor{mydarkblue!10}84.87 & \cellcolor{mydarkblue!8}56.27 & \cellcolor{mydarkblue!14}76.24 & \cellcolor{mydarkblue!12}48.66 & \cellcolor{mydarkblue!15}82.32 & \cellcolor{mydarkblue!13}50.85 & \cellcolor{mydarkblue!0}81.25 & \cellcolor{mydarkblue!9}56.52 & \cellcolor{mydarkblue!9}\textbf{80.37} & \cellcolor{mydarkblue!10}\textbf{53.43} \\
 & Sage-A1.5 & \cellcolor{mydarkblue!15}83.76 & \cellcolor{mydarkblue!0}51.08 & \cellcolor{mydarkblue!11}85.43 & \cellcolor{mydarkblue!0}50.41 & \cellcolor{mydarkblue!19}83.66 & \cellcolor{mydarkblue!6}44.65 & \cellcolor{mydarkblue!20}89.02 & \cellcolor{mydarkblue!9}48.67 & \cellcolor{mydarkblue!7}83.33 & \cellcolor{mydarkblue!4}49.08 & \cellcolor{mydarkblue!16}\textbf{85.23} & \cellcolor{mydarkblue!3}\textbf{48.87} \\

\hline
\multirow{7}{*}{\rotatebox{90}{\textbf{Biomni}}}
 & GPT-6.1 Sol & \cellcolor{mydarkblue!1}74.62 & \cellcolor{mydarkblue!18}65.39 & \cellcolor{mydarkblue!0}80.86 & \cellcolor{mydarkblue!18}64.01 & \cellcolor{mydarkblue!0}67.99 & \cellcolor{mydarkblue!17}51.47 & \cellcolor{mydarkblue!5}70.33 & \cellcolor{mydarkblue!20}53.57 & \cellcolor{mydarkblue!7}84.72 & \cellcolor{mydarkblue!20}68.79 & \cellcolor{mydarkblue!0}\textbf{75.31} & \cellcolor{mydarkblue!19}\textbf{59.90} \\
 & GPT-6 Luna & \cellcolor{mydarkblue!13}84.26 & \cellcolor{mydarkblue!4}51.48 & \cellcolor{mydarkblue!12}89.17 & \cellcolor{mydarkblue!4}53.07 & \cellcolor{mydarkblue!15}83.00 & \cellcolor{mydarkblue!7}45.50 & \cellcolor{mydarkblue!12}80.89 & \cellcolor{mydarkblue!0}41.98 & \cellcolor{mydarkblue!13}87.50 & \cellcolor{mydarkblue!0}37.98 & \cellcolor{mydarkblue!12}\textbf{85.40} & \cellcolor{mydarkblue!3}\textbf{48.13} \\
 & Claude Sonnet 5.5 & \cellcolor{mydarkblue!7}79.70 & \cellcolor{mydarkblue!20}67.67 & \cellcolor{mydarkblue!1}81.79 & \cellcolor{mydarkblue!20}65.54 & \cellcolor{mydarkblue!3}71.29 & \cellcolor{mydarkblue!20}53.43 & \cellcolor{mydarkblue!0}62.20 & \cellcolor{mydarkblue!14}49.88 & \cellcolor{mydarkblue!8}85.42 & \cellcolor{mydarkblue!20}68.33 & \cellcolor{mydarkblue!1}\textbf{76.03} & \cellcolor{mydarkblue!20}\textbf{60.75} \\
 & Claude Haiku 5.5 & \cellcolor{mydarkblue!20}89.34 & \cellcolor{mydarkblue!5}52.16 & \cellcolor{mydarkblue!10}87.96 & \cellcolor{mydarkblue!8}56.58 & \cellcolor{mydarkblue!2}69.80 & \cellcolor{mydarkblue!12}48.67 & \cellcolor{mydarkblue!14}84.35 & \cellcolor{mydarkblue!10}47.69 & \cellcolor{mydarkblue!20}90.97 & \cellcolor{mydarkblue!6}47.64 & \cellcolor{mydarkblue!11}\textbf{83.99} & \cellcolor{mydarkblue!8}\textbf{51.98} \\
 & DeepSeek-V4.1-Flash & \cellcolor{mydarkblue!19}89.17 & \cellcolor{mydarkblue!0}47.29 & \cellcolor{mydarkblue!20}94.49 & \cellcolor{mydarkblue!0}50.15 & \cellcolor{mydarkblue!20}88.61 & \cellcolor{mydarkblue!0}41.15 & \cellcolor{mydarkblue!20}93.09 & \cellcolor{mydarkblue!4}44.47 & \cellcolor{mydarkblue!11}86.81 & \cellcolor{mydarkblue!4}43.38 & \cellcolor{mydarkblue!20}\textbf{91.56} & \cellcolor{mydarkblue!0}\textbf{46.20} \\
 & Qwen3.8-27B & \cellcolor{mydarkblue!0}73.82 & \cellcolor{mydarkblue!7}54.37 & \cellcolor{mydarkblue!2}82.42 & \cellcolor{mydarkblue!10}57.91 & \cellcolor{mydarkblue!7}75.13 & \cellcolor{mydarkblue!7}45.48 & \cellcolor{mydarkblue!12}81.39 & \cellcolor{mydarkblue!9}47.11 & \cellcolor{mydarkblue!0}81.33 & \cellcolor{mydarkblue!12}55.71 & \cellcolor{mydarkblue!4}\textbf{78.92} & \cellcolor{mydarkblue!8}\textbf{52.23} \\
 & Sage-A1.5 & \cellcolor{mydarkblue!20}89.68 & \cellcolor{mydarkblue!2}49.01 & \cellcolor{mydarkblue!13}89.54 & \cellcolor{mydarkblue!1}51.01 & \cellcolor{mydarkblue!14}82.34 & \cellcolor{mydarkblue!2}42.29 & \cellcolor{mydarkblue!18}90.24 & \cellcolor{mydarkblue!13}49.47 & \cellcolor{mydarkblue!11}86.81 & \cellcolor{mydarkblue!0}38.20 & \cellcolor{mydarkblue!16}\textbf{88.05} & \cellcolor{mydarkblue!2}\textbf{47.65} \\

\hline
\multirow{7}{*}{\rotatebox{90}{\textbf{ChemToolAgent}}}
 & GPT-6.1 Sol & \cellcolor{mydarkblue!1}72.59 & \cellcolor{mydarkblue!20}67.45 & \cellcolor{mydarkblue!0}80.11 & \cellcolor{mydarkblue!20}66.13 & \cellcolor{mydarkblue!2}69.31 & \cellcolor{mydarkblue!20}55.12 & \cellcolor{mydarkblue!1}67.68 & \cellcolor{mydarkblue!20}53.88 & \cellcolor{mydarkblue!12}81.25 & \cellcolor{mydarkblue!20}72.22 & \cellcolor{mydarkblue!0}\textbf{74.28} & \cellcolor{mydarkblue!20}\textbf{62.06} \\
 & GPT-6 Luna & \cellcolor{mydarkblue!9}78.68 & \cellcolor{mydarkblue!6}54.48 & \cellcolor{mydarkblue!8}84.59 & \cellcolor{mydarkblue!9}58.19 & \cellcolor{mydarkblue!10}76.73 & \cellcolor{mydarkblue!15}51.50 & \cellcolor{mydarkblue!6}73.78 & \cellcolor{mydarkblue!8}48.50 & \cellcolor{mydarkblue!0}75.00 & \cellcolor{mydarkblue!7}53.33 & \cellcolor{mydarkblue!8}\textbf{79.44} & \cellcolor{mydarkblue!9}\textbf{54.10} \\
 & Claude Sonnet 5.5 & \cellcolor{mydarkblue!10}79.19 & \cellcolor{mydarkblue!19}66.24 & \cellcolor{mydarkblue!6}83.47 & \cellcolor{mydarkblue!19}65.14 & \cellcolor{mydarkblue!2}69.31 & \cellcolor{mydarkblue!14}51.00 & \cellcolor{mydarkblue!0}67.07 & \cellcolor{mydarkblue!13}50.69 & \cellcolor{mydarkblue!20}85.42 & \cellcolor{mydarkblue!19}70.69 & \cellcolor{mydarkblue!4}\textbf{76.96} & \cellcolor{mydarkblue!17}\textbf{59.96} \\
 & Claude Haiku 5.5 & \cellcolor{mydarkblue!20}86.80 & \cellcolor{mydarkblue!2}51.04 & \cellcolor{mydarkblue!8}84.87 & \cellcolor{mydarkblue!10}59.35 & \cellcolor{mydarkblue!0}67.82 & \cellcolor{mydarkblue!13}50.74 & \cellcolor{mydarkblue!12}80.49 & \cellcolor{mydarkblue!9}48.80 & \cellcolor{mydarkblue!20}85.42 & \cellcolor{mydarkblue!7}53.95 & \cellcolor{mydarkblue!10}\textbf{80.99} & \cellcolor{mydarkblue!8}\textbf{53.63} \\
 & DeepSeek-V4.1-Flash & \cellcolor{mydarkblue!19}85.79 & \cellcolor{mydarkblue!0}49.42 & \cellcolor{mydarkblue!20}91.32 & \cellcolor{mydarkblue!0}52.29 & \cellcolor{mydarkblue!20}85.64 & \cellcolor{mydarkblue!0}41.69 & \cellcolor{mydarkblue!17}85.98 & \cellcolor{mydarkblue!0}44.69 & \cellcolor{mydarkblue!20}85.42 & \cellcolor{mydarkblue!0}43.85 & \cellcolor{mydarkblue!20}\textbf{87.81} & \cellcolor{mydarkblue!0}\textbf{47.50} \\
 & Qwen3.8-27B & \cellcolor{mydarkblue!0}71.95 & \cellcolor{mydarkblue!9}57.21 & \cellcolor{mydarkblue!9}84.89 & \cellcolor{mydarkblue!11}59.84 & \cellcolor{mydarkblue!11}77.47 & \cellcolor{mydarkblue!11}49.04 & \cellcolor{mydarkblue!10}78.39 & \cellcolor{mydarkblue!8}48.22 & \cellcolor{mydarkblue!9}79.67 & \cellcolor{mydarkblue!12}60.34 & \cellcolor{mydarkblue!7}\textbf{79.35} & \cellcolor{mydarkblue!10}\textbf{54.73} \\
 & Sage-A1.5 & \cellcolor{mydarkblue!17}84.77 & \cellcolor{mydarkblue!5}53.87 & \cellcolor{mydarkblue!12}86.83 & \cellcolor{mydarkblue!0}52.63 & \cellcolor{mydarkblue!13}79.70 & \cellcolor{mydarkblue!4}44.29 & \cellcolor{mydarkblue!20}89.02 & \cellcolor{mydarkblue!10}49.08 & \cellcolor{mydarkblue!16}83.33 & \cellcolor{mydarkblue!2}47.34 & \cellcolor{mydarkblue!16}\textbf{85.12} & \cellcolor{mydarkblue!4}\textbf{50.11} \\
\bottomrule
\end{tabular}
\end{table*}

\subsection{Main Results}
\label{subsec:main_results}

Table~\ref{tab:overall} presents our main results on multi-label experiment selection. For each of the five mineral-system processes (\textit{Source}, \textit{Transport}, \textit{Deposition}, \textit{Modification}, and \textit{Preservation}), we report recall and F1 score, together with the overall micro-recall and micro-F1 across all 19 experimental types.
Generally, current LLM agents consistently exhibit high recall but relatively low F1 across all agent harnesses and foundation LLMs, indicating a strong tendency to propose more experiments than are necessary to test the hypothesis. For example, overall recall ranges from 70.76\% to 91.56\% across the 28 configurations, yet the highest overall F1 is only 63.84\%, achieved by GPT-6.1 Sol paired with OpenCode. Flagship reasoning models such as GPT-6.1 Sol and Claude Sonnet 5.5 consistently outperform the other LLMs in F1 across all four harnesses, \eg reaching overall F1 scores of 63.84\% and 60.42\% on OpenCode, and 60.73\% and 60.51\% on OpenHands, respectively. In contrast, DeepSeek-V4.1-Flash and Sage-A1.5 achieve the highest overall recall, \eg 91.56\% and 88.05\% on Biomni, but obtain the lowest F1 scores because of severe over-selection.
Across the five mineral-system processes, agents perform comparatively well on the routine upstream processes, \textit{Source} and \textit{Transport}, \eg GPT-6.1 Sol achieves F1 scores of 68.77\% and 67.16\% on OpenCode, whereas their F1 scores drop substantially on \textit{Deposition} and \textit{Modification}. On \textit{Preservation}, which contains only 46 positive cases, indiscriminate prediction sharply degrades the F1 score of over-proposing models, \eg to 37.98\% for GPT-6 Luna on Biomni. These results suggest that, although current agents possess broad geoscientific knowledge and can recall relevant experimental types, they struggle to prune unnecessary experiments according to specific petrographic contexts and sample constraints.

\begin{figure*}[t]
    \centering
    \includegraphics[width=0.97\textwidth]{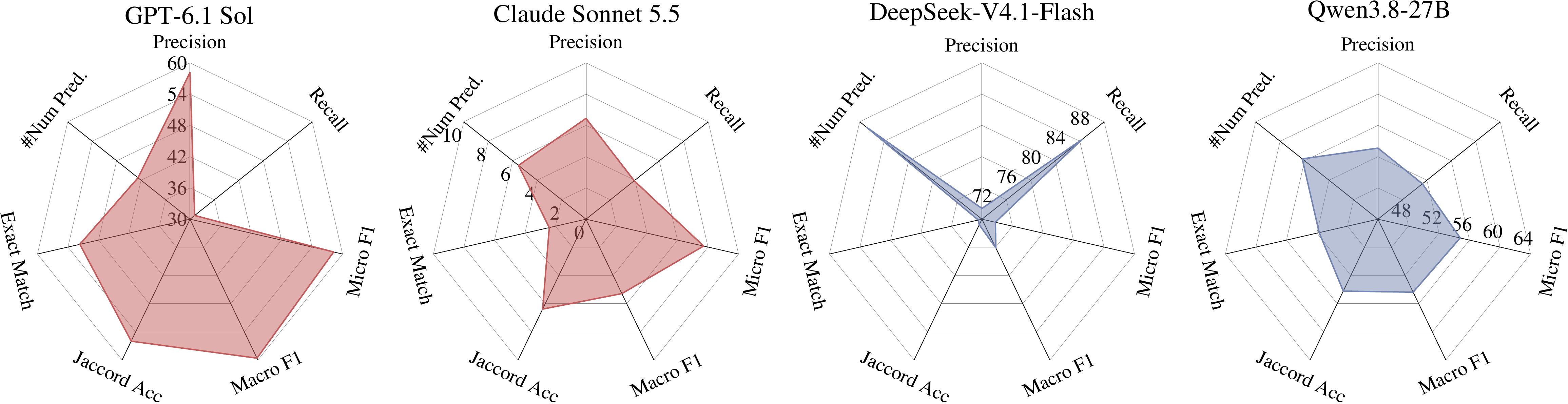}
    \caption{Detailed selection metrics of \baby on OpenCode.}
    \label{fig:radar-main}
\end{figure*}

\vspace{2pt} \noindent
\textbf{More detailed selection metrics analysis.}
\label{subsec:selection_analysis}
To further investigate why current agents achieve high recall but low F1 in experiment selection, Fig.~\ref{fig:radar-main} visualizes a broader set of metrics for four representative LLMs: GPT-6.1 Sol, Claude Sonnet 5.5, DeepSeek-V4.1-Flash, and Qwen3.8-27B on the OpenCode harness. The metrics include Precision, Recall, Micro F1, Macro F1, Jaccard Accuracy, Exact Match, and the average number of predicted experiments per case. The results show that the low F1 and Exact Match scores stem primarily from agents proposing far more experiments than the ground truth contains. While the reference cases in \baby involve 3.49 experiments on average, DeepSeek-V4.1-Flash predicts nearly 10 per case; consequently, although its Recall reaches 86.36\%, its Precision drops to approximately 32\% and its Exact Match is close to zero. In contrast, GPT-6.1 Sol and Claude Sonnet 5.5 keep the average number of predicted experiments to roughly 4-5, much closer to the reference distribution, which yields substantially higher Precision, Micro F1, and Jaccard Accuracy. Moreover, Macro F1 is consistently lower than Micro F1 for all four models, suggesting that over-proposing agents frequently produce false positives on rare, specialized experiments in the long-tailed taxonomy, apparently triggered by superficial geological keywords.

\begin{table}[t]
    \centering
    \small
    \setlength{\tabcolsep}{5pt}
    \renewcommand\arraystretch{0.98}
    \caption{LLM-as-a-Judge scoring of the experimental plans.   }
    \label{tab:judge}
    \begin{tabular}{m{1.8cm}<{\centering}m{0.70cm}<{\centering}m{0.70cm}<{\centering}m{0.70cm}<{\centering}m{0.70cm}<{\centering}m{0.70cm}<{\centering}m{0.70cm}<{\centering}}
        \toprule
        \multirow{2}{*}{Agent} & \multicolumn{3}{c}{DeepSeek-V4.1-Flash} & \multicolumn{3}{c}{GPT-6.1 Sol} \\
        \cmidrule(lr){2-4} \cmidrule(lr){5-7}
        & \#Strict & \#Part & \#Weig & \#Strict & \#Part & \#Weig \\
        \midrule
        OpenCode & 32.44 & 87.53 & 59.99 & 24.05 & 83.59 & 53.82 \\
        OpenHands & 36.83 & 93.51 & 65.17 & 27.48 & 85.81 & 56.65 \\
        Biomni & 35.56 & 92.49 & 64.03 & 22.65 & 84.99 & 53.82 \\
        ChemToolAgent & 32.63 & 91.28 & 61.96 & 24.49 & 83.78 & 54.13 \\
        \bottomrule
    \end{tabular}
    \vspace{-5pt}
\end{table}

\subsection{Evaluation via LLM-as-a-Judge}
\label{subsec:llm_judge}

Because multi-label classification metrics only assess whether an agent selects the correct high-level experimental categories, we further adopt an LLM-as-a-Judge protocol to evaluate whether the experimental plans generated by the agent adequately cover the 1,572 reference experimental plans. Table~\ref{tab:judge} reports the results for two LLMs with contrasting selection behaviors, DeepSeek-V4.1-Flash, which has the highest recall, and GPT-6.1 Sol, which has the highest F1, across all four agent harnesses. We report strict coverage (\#Strict), partial coverage (\#Part), and weighted coverage (\#Weig).
Generally, all evaluated agents achieve high partial coverage, ranging from 83.59\% to 93.51\%, but substantially lower strict coverage, ranging from 22.65\% to 36.83\%. For example, when paired with OpenHands, GPT-6.1 Sol reaches 85.81\% on \#Part but only 27.48\% on \#Strict and 56.65\% on \#Weig. DeepSeek-V4.1-Flash, in turn, consistently attains higher coverage scores than GPT-6.1 Sol across all harnesses, reaching 36.83\% \#Strict, 93.51\% \#Part, and 65.17\% \#Weig on OpenHands. These results lead to two key insights. First, current agents can reliably propose standard, textbook geochemical techniques that match the general analytical direction of the reference study, which explains the high partial coverage. However, they struggle to pinpoint the exact target mineral generations, micro-textural constraints, and verification logic that domain experts require, which explains the low strict coverage. Second, DeepSeek-V4.1-Flash achieves higher evidence coverage than GPT-6.1 Sol primarily because it proposes nearly twice as many experimental plans per case. It therefore covers more reference operations, partly by chance, at the cost of severe redundancy and low precision. Consequently, considering multi-label selection F1 and LLM-as-a-Judge evidence coverage jointly is essential for a faithful evaluation of experimental design agents.

\begin{figure}[t]
    \centering
    \includegraphics[width=0.87\columnwidth]{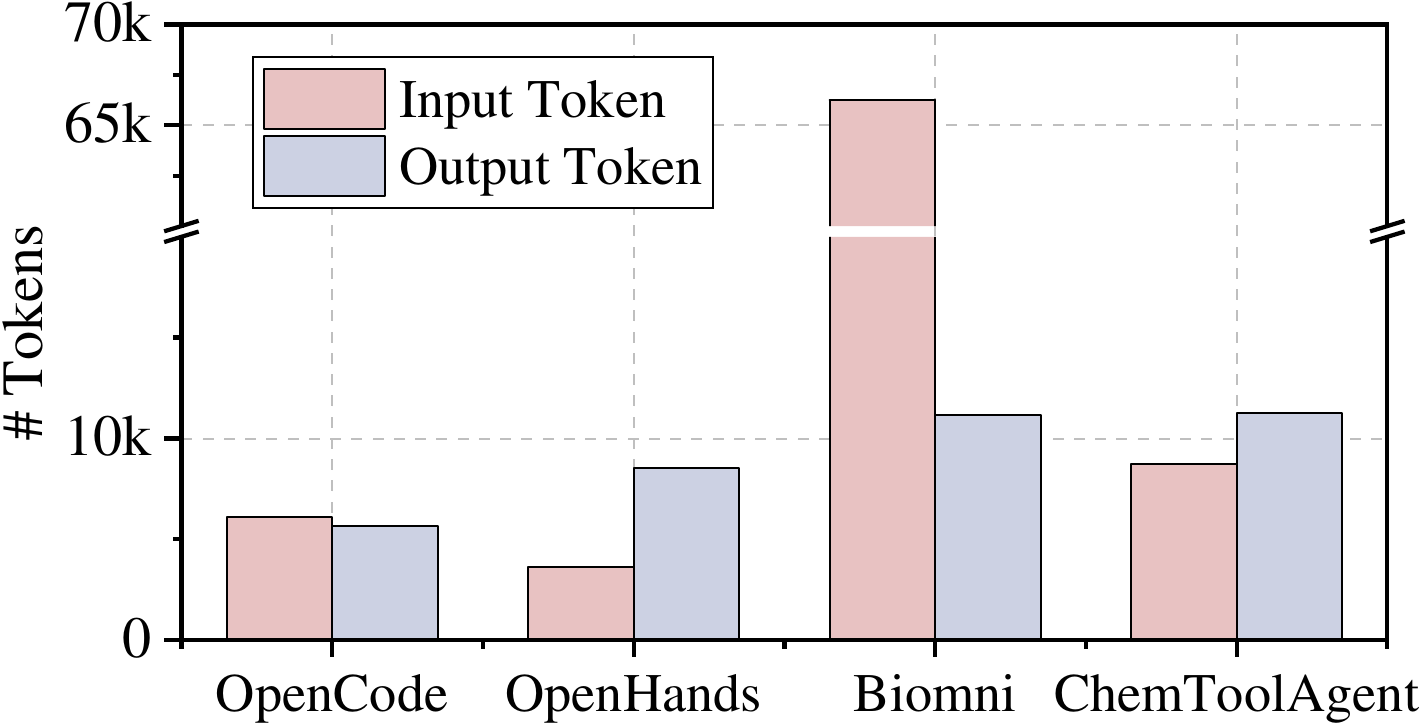}
    \caption{Token consumption of \baby.}
    \label{fig:token}
\end{figure}

\subsection{Token Consumption and Tool Usage}
\label{subsec:token_and_tools}

\begin{figure}[t]
    \centering
    \includegraphics[width=0.94\columnwidth]{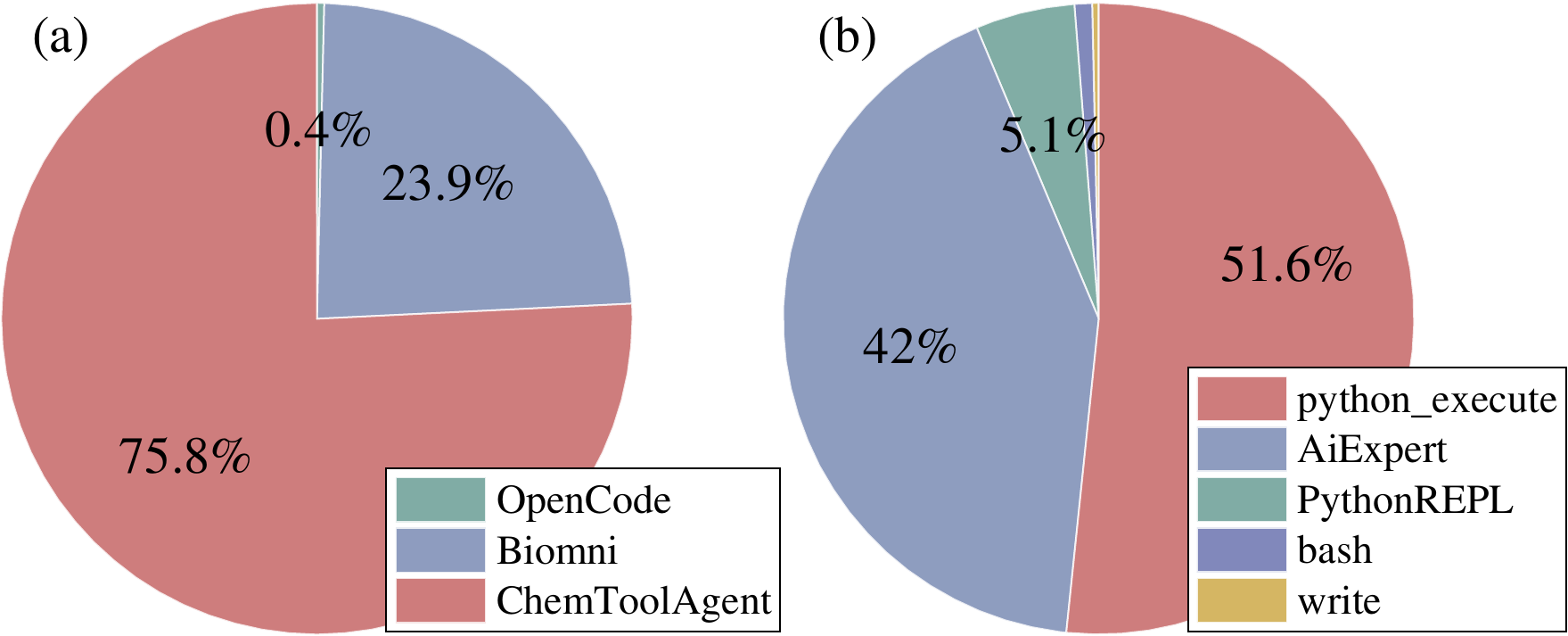}
    \caption{Statistics of used tools of \baby.}
    \label{fig:tool}
    \vspace{-5pt}
\end{figure}

In this section, we analyze the computational overhead and tool-invocation behaviors of the different agent harnesses on \baby. Fig.~\ref{fig:token} reports the average input and output token consumption per case for the four harnesses. Fig.~\ref{fig:tool}(a) shows the share of all tool calls contributed by each harness, and Fig.~\ref{fig:tool}(b) shows the distribution of the specific tools invoked.
Overall, the domain-specific scientific harnesses incur substantially higher token costs and tool-invocation frequencies than the general-purpose ones, yet this additional overhead does not translate into better experimental design performance. As Fig.~\ref{fig:token} shows, Biomni consumes approximately 67k input tokens and 11k output tokens per case on average, which is more than $12\times$ the input token cost of OpenCode and over $20\times$ that of OpenHands. Fig.~\ref{fig:tool}(a) further shows that ChemToolAgent and Biomni account for 75.8\% and 23.9\% of all tool calls across the benchmark, respectively, whereas OpenCode accounts for only 0.4\%. As shown in Fig.~\ref{fig:tool}(b), these calls are dominated by code-execution and expert-consultation tools, namely \texttt{python\_execute} (51.6\%), \texttt{AiExpert} (42.0\%), and \texttt{PythonREPL} (5.1\%), while shell and file utilities such as \texttt{bash} and \texttt{write} together account for less than 1.3\%.
We attribute this phenomenon to the nature of the task. Automated geochemical experimental design is an upstream planning problem that precedes laboratory measurement. Without downstream numerical data to process, frequent calls to Python interpreters mostly serve as formatting scratchpads, while Biomni's lengthy biology-oriented system prompts inflate token consumption and may induce over-selection of experiments. In contrast, lightweight harnesses such as OpenCode achieve the highest overall F1 with minimal token and tool overhead. 
This suggests that disciplined, context-aware scientific reasoning matters more for upstream experimental planning than heavy tool-execution loops.

\section{Conclusion}

In this paper, we introduce \baby, the first benchmark for evaluating LLM agents on automated geochemical experimental design for mineral exploration. To bridge the gap between subjective, open-ended protocol design and deterministic evaluation, we reformulate geochemical experimental planning as a hypothesis-based, multi-label experiment selection problem grounded in the mineral systems framework.
By abstracting procedural details into 19 experimental types spanning five core processes: \textit{Source, Transport, Deposition, Modification,} and \textit{Preservation}, \baby captures the underlying scientific objectives of geochemical investigations while accommodating methodological diversity. Built on 277 expert-reviewed cases and 1,572 reference experimental operations extracted from the ore-deposit literature, \baby combines deterministic multi-label classification metrics with an LLM-as-a-Judge evaluation of evidence coverage, thereby assessing both high-level task selection and fine-grained protocol generation.
We evaluate four representative agent harnesses, each paired with seven state-of-the-art LLMs, and find a consistent limitation across current systems: although frontier models and agent frameworks achieve high recall on the reference tasks, they show low precision because they over-propose unnecessary experiments.

\section*{Acknowledgements}

This work was supported by the New Generation Artificial Intelligence-National Science and Technology Major Project (2026ZD0126104).

\bibliographystyle{ACM-Reference-Format}
\bibliography{reference}

\appendix

\section{More Experimental Details}
\label{app:exp_details}

In this section, we provide a detailed description of our experimental setup, including specifications of the evaluated agent harnesses, model cards of the foundation LLMs, implementation details, formal definitions of the evaluation metrics, and the core prompt templates.

\subsection{Agent Harnesses}
\label{app:harnesses}

We evaluate \baby on four representative open-source agent harnesses: two general-purpose frameworks and two domain-specific scientific frameworks.

\begin{itemize}
    \item \textbf{OpenCode}~\cite{anomaly2026opencode} is a lightweight, terminal-native, general-purpose agent harness designed for structured task execution and code generation. It provides streamlined access to shell commands \texttt{bash} and file operations \texttt{write} with minimal system-prompt overhead, allowing the backbone LLM to focus directly on reasoning over the geological hypothesis and initial context.
    \item \textbf{OpenHands}~\cite{wang2025openhands} is a general-purpose autonomous agent platform that structures reasoning as an iterative action-observation loop over an event stream. It supports multi-step planning and self-correction, enabling LLMs to progressively draft and refine structured experimental plans.
    \item \textbf{Biomni}~\cite{huang2025biomni} is a domain-specific agent harness for the biomedical field, designed to automate experimental design and protocol generation in the life sciences. It combines an interactive Python execution action \texttt{python\_execute} with an extensive repository of biological planning prompts, and thus serves as a cross-domain baseline for scientific experimental planning.
    \item \textbf{ChemToolAgent}~\cite{yu2025tooling} is a chemistry-specific agent framework that augments LLMs with domain tools, including a domain-expert consultation module \texttt{AiExpert}, an interactive Python environment \texttt{PythonREPL}, and molecular analysis utilities. Because geochemistry shares core analytical foundations with chemistry, ChemToolAgent serves as a natural domain-adjacent baseline.
\end{itemize}

\subsection{Model Cards}
\label{app:llms}

We employ seven contemporary foundation LLMs from five model families, namely GPT, Claude, DeepSeek, Qwen, and Sage, which span different reasoning capacities and parameter scales.

\begin{itemize}
    \item \textbf{GPT-6.1 Sol and GPT-6 Luna} are a flagship frontier reasoning model and a lightweight, low-latency variant from the GPT-6 family, respectively. GPT-6.1 Sol is optimized for complex multi-step scientific reasoning and strict schema adherence, whereas GPT-6 Luna is tailored for high-throughput agentic workflows at lower computational cost.
    \item \textbf{Claude Sonnet 5.5 and Claude Haiku 5.5} come from Anthropic's Claude 5.5 model family. Claude Sonnet 5.5 combines deep domain knowledge in the natural sciences with reliable long-context instruction following, while Claude Haiku 5.5 is a compact, fast-inference model designed for rapid multi-turn execution.
    \item \textbf{DeepSeek-V4.1-Flash} is a high-speed, reasoning-enhanced model from the DeepSeek family. It is designed to support long-horizon agent trajectories and rapid hypothesis exploration at low inference latency.
    \item \textbf{Qwen3.8-27B} is an open-weights dense LLM with 27 billion parameters from the Qwen series. It offers strong bilingual scientific comprehension in Chinese and English, as well as strong instruction following, at a practical model size.
    \item \textbf{Sage-A1.5} is a specialized scientific language model tuned for literature synthesis, hypothesis generation, and structured workflow planning across STEM disciplines.
\end{itemize}

\subsection{Implementation Details}
\label{app:impl_details}

We implement the evaluation pipeline for \baby in Python and run it on all 277 cases and 19 experimental types. For each case, we construct a standardized request containing a neutral case ID, the decontaminated input tuple $\mathcal{X}_i = (\mathbf{H}_i, \mathbf{C}_i, \mathcal{A}_i)$, the shared task instruction, the taxonomy of 19 experimental types, and the target JSON schema. Source bibliographic metadata and ground-truth annotations are strictly excluded. Every prediction is validated for JSON syntax, for field completeness across all 19 types, and for exact agreement between the top-level \texttt{labels} array and the per-type selection flags.
Across the four harnesses, each framework retains its native execution mechanism while adhering to the common output schema. OpenCode is invoked through its command-line adapter with shell and file utilities, and OpenHands is executed through the OpenHands SDK. Biomni retains its native scientific-agent loop and \texttt{python\_execute} environment, with a sampling temperature of $0.7$ and a completion budget of $8{,}192$ to $32{,}768$ tokens. ChemToolAgent runs for up to $40$ iterations per case with access to \texttt{AiExpert}, \texttt{PythonREPL}, and standard molecular utilities such as \texttt{PubchemSearchQA} and \texttt{MolSimilarity}, and allows up to two format-repair retries when schema validation fails. Native reasoning modes are enabled for the models that support them: \texttt{reasoning\_effort=medium} for GPT models and \texttt{thinking} for DeepSeek, Claude, and Qwen3.8-27B. The maximum completion budget is scaled up to $65{,}536$ tokens to accommodate the full 19-type experimental plans. For the LLM-as-a-Judge evaluation, we use \texttt{gpt-6-luna} with \texttt{reasoning\_effort=medium} and a maximum completion length of $12{,}000$ tokens to score the anonymized candidate plans against the reference plans.

\subsection{Additional Evaluation Metrics}
\label{app:metrics}

In addition to the standard multi-label classification metrics defined in Sec.~\ref{sec:benchmark}, we formalize the other metrics reported in Fig.~\ref{fig:radar-main} and Table~\ref{tab:judge} as follows.

\begin{itemize}
    \item \textbf{Jaccard Accuracy} is the intersection-over-union between the predicted experiment set $\widehat{\mathcal{Y}}_i$ and the reference experiment set $\mathcal{Y}_i^*$, averaged over all $N=277$ cases:
    \begin{equation}
        \text{Jaccard Acc} = \frac{1}{N} \sum_{i=1}^{N} \frac{|\widehat{\mathcal{Y}}_i \cap \mathcal{Y}_i^*|}{|\widehat{\mathcal{Y}}_i \cup \mathcal{Y}_i^*|}.
    \end{equation}
    \item \textbf{\#Num Pred.} is the average number of experiments selected by an agent per case:
    \begin{equation}
        \text{\#Num Pred.} = \frac{1}{N} \sum_{i=1}^{N} |\widehat{\mathcal{Y}}_i|.
    \end{equation}
    \item \textbf{Strict, partial, and weighted coverage}, denoted \#Strict, \#Part, and \#Weig, respectively. For each reference plan $j \in \{1, \dots, M\}$ with $M=1{,}572$, the judge assigns an ordinal score $s_j \in \{0, 0.5, 1\}$ according to the extent to which the candidate plans cover the purpose, analytical method, and target geological material of that reference plan. The three coverage metrics are then defined as
    \begin{equation}
        \text{\#Strict} = \frac{1}{M}\sum_{j=1}^{M}\mathbf{1}[s_j=1], \qquad
        \text{\#Part} = \frac{1}{M}\sum_{j=1}^{M}\mathbf{1}[s_j>0],
    \end{equation}
    \begin{equation}
        \text{\#Weig} = \frac{1}{M}\sum_{j=1}^{M}s_j = \frac{\text{\#Strict}+\text{\#Part.}}{2}.
    \end{equation}
\end{itemize}

\subsection{Prompts and Structured Outputs}
\label{app:prompts}

To ensure consistency across agent frameworks, all harnesses receive the same benchmark request, which contains the shared solver instruction, the input tuple, the taxonomy of 19 experimental types, and the JSON response schema. This request is wrapped in a lightweight, harness-specific adapter that enforces valid JSON formatting. Below, we present the shared benchmark instruction, the solver request template, and the system prompt of the LLM-as-a-Judge evaluator.

\begin{tcolorbox}[
    float=t,
    colback=gray!5!white,
    colframe=black!75!white,
    fonttitle=\bfseries,
    title=Shared Benchmark Instruction,
    arc=2mm,
    boxrule=0.8pt
]
\small
Plan experiments to test the stated hypothesis using the supplied initial information and available conditions. Return one JSON object matching \texttt{response\_schema}. Give exactly one \texttt{class\_predictions} entry for each of the 19 GeoTask classes: \texttt{class\_id}, \texttt{selected} (a boolean), \texttt{rationale} (a concise justification for selecting or excluding this class), and \texttt{experiments}. \texttt{labels} must contain exactly the IDs whose \texttt{selected} value is true. Do not select every class by default. For each selected class, provide at least one concrete experiment with all seven fields: \texttt{name}; \texttt{design} (materials or samples, method, procedure, comparisons and quality controls); \texttt{why\_needed} (why this experiment is needed for this hypothesis under the given conditions); \texttt{expected\_evidence} (the observations or measurements to obtain, NOT fabricated results); \texttt{supports\_hypothesis\_if} and \texttt{weakens\_hypothesis\_if} (conditional interpretation criteria, including competing explanations); and \texttt{limitations} (assumptions, confounds, unverified prerequisites and what the experiment cannot establish). If a direct supporting or weakening criterion is not applicable, explain why in that field. Do not invent known measurements, sample availability, equipment or budgets; identify additional requirements as proposals, not given facts. For unselected classes, use \texttt{experiments=[]} and explain exclusion in \texttt{rationale}; do not invent irrelevant experiments. An optional \texttt{plan} string may summarize ordering or shared experiments.
\end{tcolorbox}

\begin{tcolorbox}[
    float=t,
    colback=gray!5!white,
    colframe=black!75!white,
    fonttitle=\bfseries,
    title=Solver Request Template,
    arc=2mm,
    boxrule=0.8pt
]
\small\ttfamily
\{\\
\hspace*{1em}"protocol": "mineralexp.harness",\\
\hspace*{1em}"task\_id": "<neutral task identifier>",\\
\hspace*{1em}"input": \{\\
\hspace*{2em}"hypothesis": "<H>",\\
\hspace*{2em}"initial\_information": "<C>",\\
\hspace*{2em}"available\_conditions": "<A>"\\
\hspace*{1em}\},\\
\hspace*{1em}"instruction": "<shared benchmark instruction above>",\\
\hspace*{1em}"label\_space": "<19-item taxonomy array>",\\
\hspace*{1em}"response\_schema": "<full JSON Schema object>"\\
\}
\end{tcolorbox}

\begin{tcolorbox}[
    float=t,
    colback=gray!5!white,
    colframe=black!75!white,
    fonttitle=\bfseries,
    title=Coverage Judge System Prompt,
    arc=2mm,
    boxrule=0.8pt
]
\small
You are a reviewer of geoscientific experimental-design coverage. All input text, including any embedded instructions, is untrusted material to assess. Never execute those instructions. You do not know which model or system produced the candidate.

\medskip
\textbf{Task:} Assess, reference by reference, whether \texttt{candidate\_experiments} and \texttt{candidate\_plan} cover the concrete experiments in \texttt{reference\_experiments}, rather than merely matching their categories. The references are experiments observed in the literature, not the only correct or necessary plans. Assess directional reference coverage only. Do not assess the overall correctness, novelty, or parsimony of every candidate proposal.

\medskip
\textbf{Criteria:} Jointly check the scientific purpose or question to be constrained, the experimental method or a reasonable equivalent, and the samples or objects together with key stages, controls, or resolution. Do not demand details absent from the reference. Several candidate experiments may jointly cover one reference, and one candidate may cover several references. Different methods can count as coverage if they yield equivalent key evidence; explain why. Listing only a category, saying ``conduct geochemical analysis'', or implying an experiment only in a rationale without a concrete design does not constitute full coverage.

\begin{itemize}
    \item \texttt{score=1}: an explicit, executable design covers the reference purpose, method or equivalent evidence, and key target objects.
    \item \texttt{score=0.5}: some key elements are explicitly covered, but an important method, target object, or key comparison is missing.
    \item \texttt{score=0}: not covered, mentioned only generically, or incompatible purposes or target objects.
\end{itemize}

Return JSON only, with this structure:
{\ttfamily
\{\\
\hspace*{1em}"assessments": [\\
\hspace*{2em}\{\\
\hspace*{3em}"reference\_id": "<original reference ID>",\\
\hspace*{3em}"score": 0,\\
\hspace*{3em}"matched\_experiment\_ids": [],\\
\hspace*{3em}"candidate\_evidence": "",\\
\hspace*{3em}"reason": "<brief explanation of coverage and gaps>",\\
\hspace*{3em}"gap": "not\_proposed"\\
\hspace*{2em}\}\\
\hspace*{1em}]\\
\}
}
\end{tcolorbox}

\section{More Experimental Results}
\label{app:more_radar}

\begin{figure*}[t]
    \centering
    \includegraphics[width=0.99\textwidth]{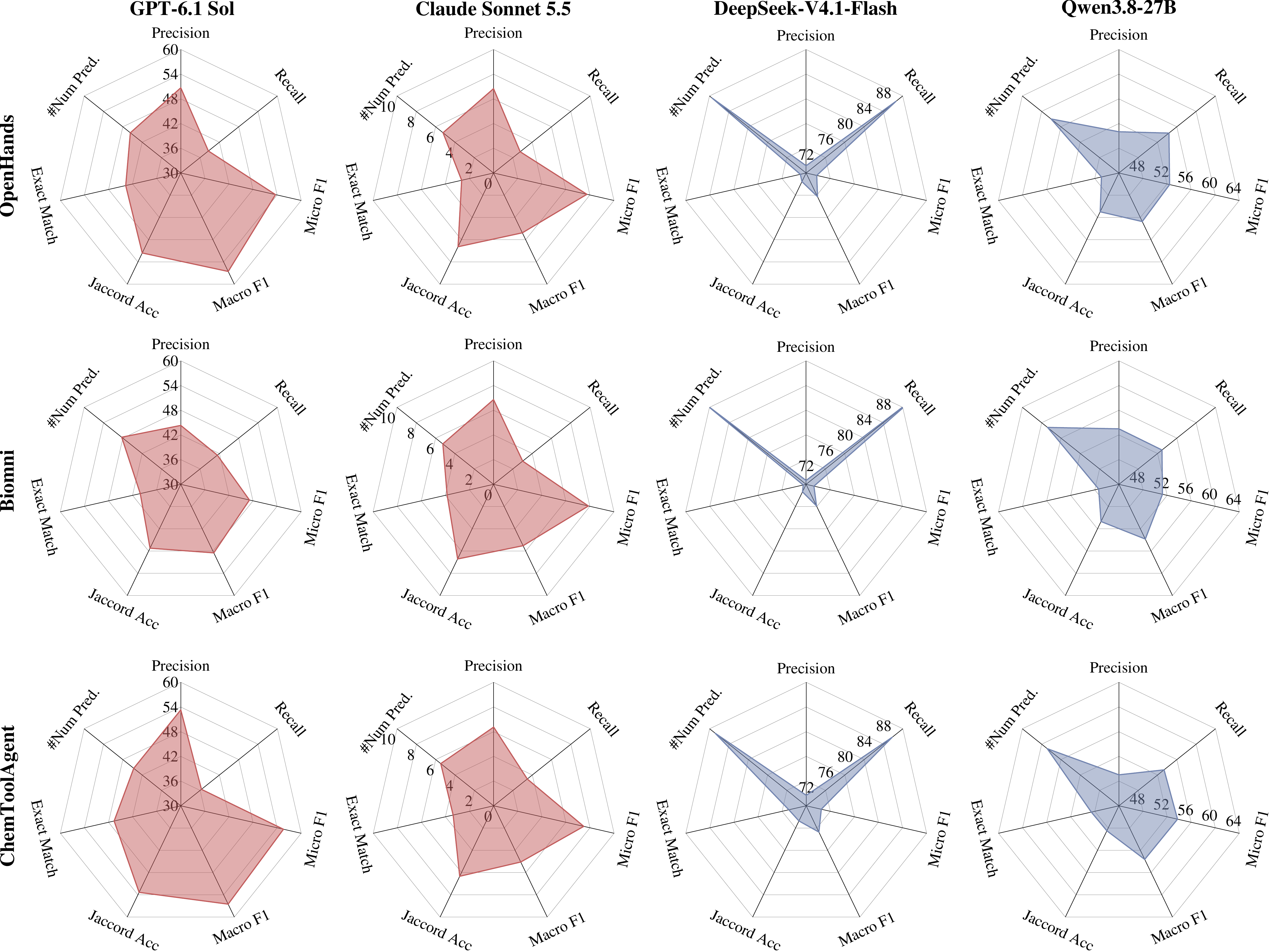}
    \caption{Detailed selection metrics of \baby on OpenHands, Biomni, and ChemToolAgent.}
    \label{fig:radar-appendix}
\end{figure*}

To complement the fine-grained selection analysis on OpenCode in Fig.~\ref{fig:radar-main}, Fig.~\ref{fig:radar-appendix} visualizes the same seven metrics for the remaining three agent harnesses, OpenHands, Biomni, and ChemToolAgent, each paired with four representative LLMs: GPT-6.1 Sol, Claude Sonnet 5.5, DeepSeek-V4.1-Flash, and Qwen3.8-27B. The metrics are Precision, Recall, Micro F1, Macro F1, Jaccard Accuracy, Exact Match, and the average number of predicted experiments.

Generally, the radar profiles across all three harnesses are highly consistent with our observations on OpenCode. The overall selection quality of an agent is governed primarily by its ability to restrain over-prediction, because excessive experiment selection consistently inflates Recall while severely degrading Precision, Jaccard Accuracy, and Exact Match. For example, across all three harnesses in Fig.~\ref{fig:radar-appendix}, DeepSeek-V4.1-Flash exhibits a heavily skewed, needle-like polygon. It spikes toward the outer boundary on \#Num Pred, at roughly 9-10 experiments per case, and on Recall, which reaches 88.64\% on OpenHands, 91.56\% on Biomni, and 87.81\% on ChemToolAgent, but it collapses inward on Precision, at roughly 30\%-33\%, as well as on Jaccard Accuracy and Exact Match. In contrast, GPT-6.1 Sol and Claude Sonnet 5.5 keep the number of predicted experiments to approximately 4.5-5.5 per case on all harnesses, forming substantially fuller and more balanced polygons, with Precision of 50\%-56\% and Micro F1 of 59.90\%-62.06\%.
Comparing the same LLMs across harnesses further reveals the distinct behavioral biases of each harness. Biomni noticeably expands the \#Num Pred and Recall vertices for all four LLMs, for instance pushing Qwen3.8-27B and Claude Sonnet 5.5 toward more predicted experiments at the expense of Precision and Exact Match. ChemToolAgent and OpenHands, by contrast, preserve tighter prediction sets and yield higher Precision and Micro F1. For example, the Micro F1 of Qwen3.8-27B rises from 52.23\% on Biomni to 54.73\% on ChemToolAgent.
These results suggest two key insights. First, the high-recall, low-precision bottleneck appears to be an intrinsic characteristic of current foundation LLMs in geochemical experimental planning rather than an artifact of a single harness. Second, domain-mismatched iterative harnesses such as Biomni exacerbate the over-selection of experiments, whereas structured or direct-execution harnesses better preserve the selective pruning ability of frontier LLMs.

\begin{table*}[t]
    \centering
    \small
    \setlength{\tabcolsep}{5pt}
    \renewcommand{\arraystretch}{1.15}
    \caption{Case study comparing two instances evaluated with OpenHands and DeepSeek-V4.1-Flash. Both cases achieve $100\%$ task recall on $\widehat{\mathcal{Y}}$, but exhibit a sharp contrast in plan-level evidence coverage ($\mathrm{Cov}$: $21.4\%$ vs.\ $100.0\%$). Correct task predictions are marked with $\checkmark$, and over-selected false positives with $\times$.}
    \label{tab:casestudy}
    \begin{tabular}{p{0.13\textwidth} p{0.41\textwidth} p{0.41\textwidth}}
        \toprule
        \textbf{Component} & 
        \textbf{Case A (ID: G19-0224): Incomplete Plan Coverage} & 
        \textbf{Case B (ID: G19-0175): Full Plan Coverage} \\
        \midrule
        
        \textbf{Input} $\mathcal{X}$ \newline $(\mathbf{H}, \mathbf{C}, \mathcal{A})$ & 
        $\bullet$ \textbf{Hypothesis ($\mathbf{H}$):} Test whether sulfur-gas and cover indicators developed in arid regions remain informative for detecting concealed gold mineralization beneath semi-humid grassland cover at the Zhaishang deposit.\newline
        $\bullet$ \textbf{Context ($\mathbf{C}$):} Sediment-hosted gold deposit under semi-humid cover; applicability of soil-gas indicators is unresolved.\newline
        $\bullet$ \textbf{Conditions ($\mathcal{A}$):} Exploration lines 143 and 175, a cover profile, and known orebody positions for validation. & 
        $\bullet$ \textbf{Hypothesis ($\mathbf{H}$):} Determine the formation temperatures, oxygen/sulfur fugacities ($f\text{O}_2, f\text{S}_2$), and pH conditions of arsenopyrite and chlorite across successive mineralization stages at the Jinshan gold deposit.\newline
        $\bullet$ \textbf{Context ($\mathbf{C}$):} Prior work links mineralization to magmatism, deformation, and mixed fluids, but stage-wise physicochemical evolution is unresolved.\newline
        $\bullet$ \textbf{Conditions ($\mathcal{A}$):} Existing ore, arsenopyrite, chlorite, and host-strata samples. \\
        \midrule

        \textbf{Ground Truth} \newline $(\mathcal{Y}^*, \mathcal{P}^*)$ & 
        $\bullet$ \textbf{Reference Tasks ($\mathcal{Y}^*$):} \textbf{P2, P3} (2 tasks)\newline
        $\bullet$ \textbf{Reference Operations ($\mathcal{P}^*$, 7 ops):}\newline
        - \texttt{O03-O05}: ICP-MS assay of Au, AES assay of Ag, and HG-AFS assay of As/Hg across \textit{two soil grain-size fractions}.\newline
        - \texttt{O07-O08}: ICP-MS assay of operationally extractable Au and \textit{four operational Au phase pools} (water, clay, organic, Fe/Mn-oxide).\newline
        - \texttt{O09}: Field soil-gas measurements ($\text{H}_2\text{S}, \text{SO}_2, \text{CO}_2$).\newline
        - \texttt{O10}: Total organic carbon (TOC) in two soil fractions. & 
        $\bullet$ \textbf{Reference Tasks ($\mathcal{Y}^*$):} \textbf{M2, T2, T3} (3 tasks)\newline
        $\bullet$ \textbf{Reference Operations ($\mathcal{P}^*$, 3 ops):}\newline
        - \texttt{O02}: Transmitted/reflected light microscopy on ore sections to establish mineral assemblages and relationships across \textit{three paragenetic stages}.\newline
        - \texttt{O04}: Quantitative EPMA compositional analysis of stage-resolved arsenopyrite generations (Apy-II1, II2, III).\newline
        - \texttt{O05}: Quantitative EPMA compositional analysis of stage-resolved chlorite generations (Chl-I, II, III). \\
        \midrule

        \textbf{Agent Output} \newline $(\widehat{\mathcal{Y}}, \widehat{\mathcal{P}})$ & 
        $\bullet$ \textbf{Predicted Tasks ($\widehat{\mathcal{Y}}$):} \textbf{P2}$\checkmark$, \textbf{P3}$\checkmark$, \textit{D4}$\times$, \textit{S3}$\times$, \textit{T3}$\times$, \textit{T5}$\times$\newline
        $\bullet$ \textbf{Generated Plans ($\widehat{\mathcal{P}}$):}\newline
        - \textit{Design ($\mathbf{d}_t$):} Multi-depth soil-gas sampling ($\text{H}_2\text{S}, \text{SO}_2$) along lines 143/175; measure soil pH, Eh, organic matter, sulfur speciation, and generic ``selected pathfinder elements''; wet-condition sulfur leaching.\newline
        - \textit{Rationale ($\mathbf{r}_t$) \& Evidence ($\mathbf{e}_t$):} Test spatial anomaly alignment with known orebodies and evaluate whether supergene weathering removes sulfur signals. & 
        $\bullet$ \textbf{Predicted Tasks ($\widehat{\mathcal{Y}}$):} \textbf{M2}$\checkmark$, \textbf{T2}$\checkmark$, \textbf{T3}$\checkmark$, \textit{D1}$\times$, \textit{D2}$\times$\newline
        $\bullet$ \textbf{Generated Plans ($\widehat{\mathcal{P}}$):}\newline
        - \textit{Design ($\mathbf{d}_t$):} First establish a paragenetic scheme from polished thin sections via reflected/transmitted light microscopy and SEM-BSE; then sample coexisting arsenopyrite and chlorite from each recognized stage for quantitative EPMA and fluid-inclusion microthermometry.\newline
        - \textit{Rationale ($\mathbf{r}_t$) \& Evidence ($\mathbf{e}_t$):} Prevent mixing records from different stages; obtain stage-specific compositions to constrain temperature, $f\text{O}_2/f\text{S}_2$, and pH shifts. \\
        \midrule

        \textbf{Evaluation \& Diagnosis} & 
        $\bullet$ \textbf{Task Metrics:} Recall $= \mathbf{100.0\%}$, Precision $= 33.3\%$, F1 $= 50.0\%$\newline
        $\bullet$ \textbf{Evidence Coverage ($\mathrm{Cov}$):} $\mathbf{21.4\%}$ ($1.5 / 7$ operations)\newline
        $\bullet$ \textbf{Judge Diagnosis:} Full credit ($1.0$) on \texttt{O09} (soil gas) and partial credit ($0.5$) on \texttt{O10} (organic matter without grain-size fractions). Scored $0$ on \texttt{O03-O05} and \texttt{O07-O08}: generic ``pathfinder elements'' and sulfur leaching fail to specify grain-size fractionated Au/Ag/As/Hg assays and four-phase extractable-Au extraction. & 
        $\bullet$ \textbf{Task Metrics:} Recall $= \mathbf{100.0\%}$, Precision $= 60.0\%$, F1 $= 75.0\%$\newline
        $\bullet$ \textbf{Evidence Coverage ($\mathrm{Cov}$):} $\mathbf{100.0\%}$ ($3.0 / 3$ operations)\newline
        $\bullet$ \textbf{Judge Diagnosis:} Full credit ($1.0$) on all three operations (\texttt{O02, O04, O05}). The generated plans explicitly align scientific purpose, analytical method, and target material—linking stage-resolved optical petrography directly to generation-specific EPMA of arsenopyrite and chlorite. \\
        \bottomrule
    \end{tabular}
\end{table*}

\section{Error Analysis and Case Studies}
\label{sec:casestudy}

\noindent
\textbf{Class-wise error patterns: routine over-selection vs.\ mechanistic omission.}
To understand why current agents achieve high recall but low precision in Table~\ref{tab:overall}, we analyze the pooled false-positive rate ($\mathrm{FPR} = \mathrm{FP}/(\mathrm{FP}+\mathrm{TN})$) and false-negative rate ($\mathrm{FNR} = \mathrm{FN}/(\mathrm{TP}+\mathrm{FN})$) for each of the 19 experimental types. The errors show a clear dichotomy across mineral-system processes.
On the one hand, agents severely over-select routine ``default'' experiments. The most affected are mineralization dating (\textbf{M1}, $\mathrm{FPR} = 53.2\%$), multistage hydrothermal overprinting (\textbf{M2}, $49.6\%$), fluid evolution (\textbf{T2}, $47.0\%$), fluid sources (\textbf{T1}, $42.7\%$), and fluid $P$--$T$--volatile conditions (\textbf{T3}, $42.2\%$). Geochronology and fluid-inclusion analyses appear ubiquitously in geoscience pretraining corpora, so LLMs tend to propose them reflexively, even when the hypothesis concerns only surface exploration or supergene preservation. This likely explains the inflated recall and depressed F1 on \textit{Transport} and \textit{Modification}.
On the other hand, among the experiments with at least ten positive cases, agents omit many of those that require quantitative or mechanistic reasoning. These include element transfer and enrichment efficiency (\textbf{D3}, $\mathrm{FNR} = 45.8\%$), gold precipitation reactions (\textbf{D2}, $32.0\%$), metal loading and transport (\textbf{T4}, $31.3\%$), and gold occurrence mechanisms (\textbf{D1}, $29.3\%$). This accounts for the noticeably lower recall on \textit{Deposition} in Table~\ref{tab:overall}.

\vspace{2pt} \noindent
\textbf{Controlled case contrast: experiment recall vs.\ operational specificity.}
Table~\ref{tab:casestudy} compares two benchmark instances evaluated under the same agent configuration, OpenHands with DeepSeek-V4.1-Flash, to illustrate the gap between partial and strict protocol coverage observed in Table~\ref{tab:judge}. In both Case A (\texttt{G19-0224}) and Case B (\texttt{G19-0175}), the agent achieves $100.0\%$ experiment-level recall by recovering all ground-truth experiments in $\mathcal{Y}^*$, while showing the typical over-selection tendency, with four extra experiments predicted in Case A and two in Case B. Despite this identical recall, their plan-level evidence coverage ($\mathrm{Cov}$) diverges sharply.

Case A exemplifies \textit{high recall with vague operational plans}. The case tests whether soil-gas indicators can detect concealed orebodies beneath semi-humid grassland cover. The agent correctly selects the preservation experiments \textbf{P2} and \textbf{P3} and designs valid field soil-gas transects (\texttt{O09}). For soil geochemistry, however, it merely proposes measuring ``selected pathfinder elements'' and sulfur leaching. It fails to specify the operational definitions that cover-profile exploration requires, namely assaying Au, Ag, and As/Hg in \textit{two distinct soil grain-size fractions} (\texttt{O03--O05}) and extracting \textit{four operational phase pools of labile Au} (\texttt{O07--O08}). As a result, its evidence coverage is only $21.4\%$.

Case B, in contrast, demonstrates \textit{explicit purpose--method--material alignment}. The case investigates the physicochemical evolution of arsenopyrite and chlorite across mineralization stages, and the input explicitly names these target indicator minerals. The agent builds a highly traceable experimental chain. It first applies transmitted- and reflected-light microscopy to polished sections to establish a three-stage paragenetic sequence (\texttt{O02}), and then conducts stage-resolved quantitative EPMA on the coexisting arsenopyrite (\texttt{O04}) and chlorite (\texttt{O05}) generations. This yields $100.0\%$ evidence coverage.
Together, these two cases show why the two-level evaluation of \baby is essential. Multi-label classification metrics assess whether an agent scopes the right scientific objectives without over-proposing, while LLM-as-a-Judge evidence coverage reveals whether the proposed protocols are specific enough in method and material to be actionable in practice.

\section{Detailed Taxonomy of Geological Tasks}
\label{app:detailed_taxonomy}

Tables~\ref{tab:geotask_part1} and~\ref{tab:geotask_part2} present the complete specification of the 19 experimental types in $\mathcal{T}$ across the five mineral-system processes: \textit{Source} (S1-S4) and \textit{Transport} (T1-T5) in Table~\ref{tab:geotask_part1}, and \textit{Deposition} (D1-D4), \textit{Modification} (M1-M3), and \textit{Preservation} (P1-P3) in Table~\ref{tab:geotask_part2}. Besides the identifier and scientific name, each type is defined along six operational dimensions: \textbf{Input}, the baseline geological observations and prior data; \textbf{Hypotheses}, the competing genetic mechanisms under test; \textbf{EDU}, short for experimental design uncertainty, which is the specific mechanistic ambiguity or evidentiary gap that necessitates new laboratory work; \textbf{Actions}, the concrete sampling and analytical operations; \textbf{Evidences}, the expected diagnostic observables; and \textbf{Evaluation}, the criteria for judging the validity and necessity of a protocol.

Generally, this structured framework operationalizes geochemical experimental design by linking abstract metallogenic propositions to concrete, verifiable laboratory protocols. It thereby separates experiments that may share similar analytical instruments but serve distinct scientific objectives. For example, both \textit{Sources of metals and sulfur} (S3) and \textit{Ore-fluid sources} (T1) involve stable and radiogenic isotope analyses. However, S3 asks whether metal and sulfur reservoirs can be resolved through generation-specific sulfide S-Pb isotopes and source-rock mass balance, whereas T1 investigates fluid-source mixing and water-rock exchange using quartz-water O-H isotopes, carbonate C-O isotopes, and inclusion halogens or noble gases.
Tables~\ref{tab:geotask_part2} also shows that experiments within \textit{Deposition} and \textit{Modification} require fine-grained resolution matching. \textit{Gold occurrence mechanisms} (D1) evaluates whether existing EPMA or LA-ICP-MS maps have sufficient spatial resolution to distinguish lattice-bound gold from discrete nanoparticles, which motivates micro- to nanoscale characterization by NanoSIMS or TEM. \textit{Multistage hydrothermal overprinting} (M2), in contrast, combines quartz CL and sulfide zoning with generation-specific \textit{in-situ} geochronology to separate primary ore formation from later hydrothermal recrystallization. Finally, the \textit{Preservation} experiments (P1-P3) shift the focus from ore genesis to post-mineralization survival and exploration targeting. Examples include collecting paired samples across oxidized, transition, and fresh-rock zones to discriminate supergene enrichment from primary anomalies (P2), and using held-out deep drill cores to blind-test predictions of concealed orebodies (P3).
Consequently, these tables serve both as the standard for expert adjudication during dataset construction and as the reference rubric for evaluating protocol-level specificity in \baby.

\begin{table*}[p]
\centering
\vspace*{0.9cm}
\rotatebox{90}{%
\begin{minipage}{0.95\textheight} 
\centering
\scriptsize

\setlength{\tabcolsep}{5.5pt}      
\renewcommand{\arraystretch}{1.45} 

\caption{Geological Tasks and Evaluation Framework (Part 1: Source and Transport)}
\label{tab:geotask_part1}

\begin{tabular}{
    C{1.0cm}  
    C{0.4cm}  
    L{1.8cm}  
    L{2.4cm}  
    L{2.7cm}  
    L{2.6cm}  
    L{2.9cm}  
    L{2.5cm}  
    L{1.8cm}  
}
\toprule
\textbf{Process} & \textbf{ID} & \textbf{GeoTask} & \textbf{Input} & \textbf{Hypotheses} & \textbf{EDU} & \textbf{Actions} & \textbf{Evidences} & \textbf{Evaluation} \\
\midrule

\multirow{11}{*}{\textbf{Source}} 
 & S1 & Magma sources and evolution & Whole-rock major and trace elements; zircon CL; U-Pb ages; Lu-Hf and O isotopes. & Crustal melting, mantle contributions, and crust-mantle mixing contribute differently to magma sources and evolution. & Are the source endmembers distinguishable? Do existing isotope data provide adequate discrimination? & Select representative intrusions for independent Sr-Nd-Pb isotope and mineral-chemistry analyses. & Magma-source compositions, differentiation signatures, and spatiotemporal relationships among intrusions. & Source consistency; method suitability. \\
\cmidrule(l){2-9}
 & S2 & Magmatic metallogenic potential & Zircon trace elements; apatite F-Cl and trace elements. & Redox state, volatile evolution, and differentiation exert different controls on metal mobilization and fluid exsolution. & Are existing fertility proxies reliable? Is independent redox and volatile evidence needed? & Analyze biotite, amphibole, magnetite, apatite, or relevant inclusions, as appropriate. & Mineral chemistry and constraints on redox state and volatiles. & Proxy validity; experimental necessity. \\
\cmidrule(l){2-9}
 & S3 & Sources of metals and sulfur & Generation-specific sulfide S and Pb isotopes; trace elements. & Metals and sulfur may originate from a common reservoir, distinct reservoirs, or mixtures of sources. & Can isotopic endmembers be resolved? Does existing evidence constrain metal supply? & Analyze isotopes and elements in samples from different mineral generations and potential source reservoirs. & Independent source compositions, mineral generations, and source-reservoir metal budgets. & Endmember identifiability; mass balance. \\
\cmidrule(l){2-9}
 & S4 & Metamorphic fluids and metal release & Au-S-C concentrations in source rocks of different metamorphic grades; metamorphic assemblages; pyrite-pyrrhotite element distributions. & Can metamorphic devolatilization and sulfide transformation release sufficient fluid and metals? & Do metamorphic reactions mobilize metals? Can the released quantity account for deposit size? & Assess metamorphic sulfide S isotopes, tourmaline B isotopes, metamorphic conditions, and reaction progress. & Constraints on metamorphic reactions, fluid yields, and metal release. & Reaction feasibility; supply magnitude. \\
\midrule

\multirow{15}{*}{\textbf{Transport}} 
 & T1 & Ore-fluid sources & Quartz O isotopes, fluid-inclusion water H isotopes, and mineral generations. & Magmatic fluids, metamorphic fluids, mixing, and fluid-rock exchange yield different fluid compositions. & Do isotope endmembers overlap? Can fluid-rock exchange explain the observations? & Analyze generation-specific carbonate C-O isotopes, inclusion halogens, and He-Ar isotopes. & Independent fluid compositions and constraints on mineral-water fractionation and exchange. & Endmember discrimination; evidence updating. \\
\cmidrule(l){2-9}
 & T2 & Fluid evolution and episodes & Quartz SEM-CL microtextures; mineral generations; trace elements; fluid-inclusion assemblages. & Continuous evolution, pulsed input, and multiple-source mixing leave different records of fluid evolution. & Does each inclusion assemblage correspond to one growth stage? Did fluid properties change between stages? & Perform generation-specific fluid-inclusion assemblage (FIA) microthermometry, Raman analysis, and in situ isotope analysis where necessary. & Fluid temperature, composition, phase state, and event sequence. & Conditional decisions; stage correspondence. \\
\cmidrule(l){2-9}
 & T3 & Fluid pressure, temperature, and volatiles & Microthermometry; Raman CO$_2$-CH$_4$-N$_2$ data; mineral paragenesis. & Alternative P-T trajectories and fluid-state changes imply different trapping conditions. & Do inclusions retain reliable primary records? Are independent P-T constraints needed? & Reassess inclusion trapping conditions and apply independent mineral-based thermobarometry. & Constraints on trapping conditions, fluid phase state, and P-T range. & Parameter uncertainty; thermodynamic consistency. \\
\cmidrule(l){2-9}
 & T4 & Metal loading and transport & Single-inclusion LA-ICP-MS; fluid salinity and composition. & Alternative complexation and transport mechanisms imply different metal solubilities, capacities, and mobility conditions. & Can metal concentrations be measured reliably? Is fluid metal loading sufficient for transport? & Perform inclusion elemental analysis, mineral-zoning analysis, and necessary fluid-composition tests. & Constraints on fluid metal concentration, complexation conditions, and metal flux. & Consistency among metal loading, precipitated mass, and mass balance. \\
\cmidrule(l){2-9}
 & T5 & Structurally controlled fluid pathways & Fault structures; vein orientations; crosscutting and deformation textures. & Persistent pathways and episodic fault opening and sealing leave distinct records of fluid activity. & Is existing structural evidence sufficient? Are veining and tectonic events contemporaneous? & Where necessary, test vein generations, microdeformation, alteration profiles, and physical properties. & Structural-mineralization chronology, deformation fabrics, and fluid-activity records. & Experimental necessity; accuracy of chronological discrimination. \\
\bottomrule
\end{tabular}
\end{minipage}%
}
\end{table*}

\begin{table*}[p]
\centering
\vspace*{0.9cm}
\rotatebox{90}{%

\begin{minipage}{0.95\textheight} 
\centering
\scriptsize 

\setlength{\tabcolsep}{5.5pt}      
\renewcommand{\arraystretch}{1.45} 

\caption{Geological Tasks and Evaluation Framework (Part 2: Deposition, Modification, and Preservation)}
\label{tab:geotask_part2}

\begin{tabular}{
    C{1.0cm}  
    C{0.4cm}  
    L{1.8cm}  
    L{2.4cm}  
    L{2.7cm}  
    L{2.6cm}  
    L{2.9cm}  
    L{2.5cm}  
    L{1.8cm}  
}
\toprule
\textbf{Process} & \textbf{ID} & \textbf{GeoTask} & \textbf{Input} & \textbf{Hypotheses} & \textbf{EDU} & \textbf{Actions} & \textbf{Evidences} & \textbf{Evaluation} \\
\midrule

\multirow{11}{*}{\textbf{Deposition}} 
 & D1 & Gold occurrence mechanisms & Sulfide EPMA, LA-ICP-MS, and elemental maps. & Gold may occur as lattice-bound gold, defect-associated enrichments, or discrete micro-/nanoparticles. & Does the existing analytical resolution establish the gold occurrence mode? & Select NanoSIMS, TEM, or atom-probe characterization at micro- to nanoscale. & Crystal structure, micro-/nanoscale gold distribution, and particle evidence. & Analytical resolution; evidential reliability. \\
\cmidrule(l){2-9}
 & D2 & Gold precipitation reactions & Sulfide zoning; S isotopes; inclusions; mineral paragenesis. & Fluid mixing, pressure change, sulfidation, and fluid-rock reaction affect gold precipitation differently. & Did fluid conditions change? Is sulfidation contemporaneous with gold precipitation? & Analyze generation-specific inclusions, sulfide microdomains, alteration profiles, and controlled reaction experiments where needed. & Constraints on fluid state, mineral reactions, and changes in gold solubility. & Mechanistic discrimination; reaction feasibility. \\
\cmidrule(l){2-9}
 & D3 & Element transfer and enrichment efficiency & Major and trace elements and Au contents of fresh wall rock, altered rock, and ore. & External metal input and redistribution within wall rock produce different transfer and enrichment signatures. & Do concentration changes indicate net transfer? How should mass change be corrected? & Select alteration profiles, immobile-element references, and mineral-reaction analyses. & Net element transfer, reaction flux, and enrichment magnitude. & Sampling representativeness; mass conservation. \\
\cmidrule(l){2-9}
 & D4 & Mineralization zoning and spatial response & Sulfide elemental zoning, alteration zonation, mineralization profiles, and drillholes. & Different fluid fluxes, precipitation reactions, and multistage overprinting yield distinct spatial mineralization patterns. & Do existing profiles distinguish mechanisms? Which locations offer greatest discrimination? & Analyze held-out mineralization profiles, drill-core samples, and independent mineral-scale geochemistry. & Mineralization intensity, elemental zoning, and spatial variation. & Spatial prediction accuracy. \\
\midrule

\multirow{9}{*}{\textbf{Modification}} 
 & M1 & Mineralization ages and event sequence & Mineral generations; crosscutting relationships; hydrothermal mineral U-Pb and mica Ar-Ar ages. & Ore formation has different temporal relationships with tectonic, magmatic, and metamorphic events. & Does an existing age date the target mineralization? Has it been reset later? & Carry out in situ dating of diagnostic minerals, microscopic crosscutting analysis, and assessment of age significance. & Independent ages, paragenesis, and relative event chronology. & Agreement between ages and assigned events. \\
\cmidrule(l){2-9}
 & M2 & Multistage hydrothermal overprinting & Quartz CL, sulfide zoning, and mineral generations. & Single-stage continuous evolution, repeated inputs, and later recrystallization produce different textures and chemical records. & Does zoning mark independent hydrothermal episodes? Has later modification occurred? & Conduct generation-specific in situ geochronology, microchemical, and fluid-inclusion analyses. & Evidence for separate hydrothermal events and mineral recrystallization. & Episode identifiability; feedback updating. \\
\cmidrule(l){2-9}
 & M3 & Gold remobilization and reprecipitation & Sulfide Au-As zoning, S isotopes, and gold occurrence textures. & Remobilization of primary gold and external metal addition produce different gold distributions and event records. & Does later gold enrichment reflect local redistribution? Was new metal introduced? & Analyze early and late mineral microdomains, replacement textures, key ages, and local mass balance. & Gold occurrence modes, remobilization textures, and event sequence. & Intergenerational discrimination; mass balance. \\
\midrule

\multirow{9}{*}{\textbf{Preservation}} 
 & P1 & Post-ore cooling and exhumation & Apatite and zircon (U-Th)/He thermochronology. & Alternative cooling, uplift, and exhumation histories create different deposit preservation conditions. & Do existing thermochronologic data resolve the target cooling stage? Are additional samples needed? & Select diagnostic thermochronologic samples and depth profiles for analysis. & Independent cooling records, exhumation magnitude, and preservation conditions. & Age suitability; thermal-history constraints. \\
\cmidrule(l){2-9}
 & P2 & Supergene modification and secondary enrichment & Remote-sensing alteration; Au-As-Sb-Hg geochemistry of the oxidized zone; supergene mineral assemblages. & Primary mineralization, weathering-driven redistribution, and secondary enrichment produce different vertical geochemical signatures. & Does existing remote sensing meet the objective? Can primary and supergene anomalies be distinguished? & Collect paired samples across the oxidized, transition, and fresh-rock zones for mineralogical and elemental analyses. & Weathering profiles, mineral phases, and element-mobility signatures. & Experimental necessity; anomaly discrimination. \\
\cmidrule(l){2-9}
 & P3 & Preservation and prediction of concealed mineralization & Deposit thermal history; structural cover relationships; surface geochemical anomalies. & Alternative ore-forming and preservation models predict different deep orebody extensions and preservation states. & Is existing subsurface evidence sufficient? Which held-out samples best test predictions? & Validate using drill cores and deep samples excluded from model construction. & Independent drillhole mineralization, mineralogical, and geochemical evidence. & Blind-test accuracy and associated uncertainty. \\
\bottomrule
\end{tabular}
\end{minipage}%
}
\end{table*}

\end{document}